\documentclass[letterpage, 11pt, notitlepage]{article}

\usepackage[margin=1.0in]{geometry}

\usepackage{cite}
\usepackage{amsmath,amssymb,amsfonts}
\usepackage{algorithmic}
\usepackage{graphicx}
\usepackage{textcomp}

\usepackage{hyperref}       
\usepackage{url}            
\usepackage{booktabs}       
\usepackage{nicefrac}       
\usepackage{microtype}      
\usepackage{bbold}
\usepackage{capt-of}

\usepackage{amsthm, epsfig,epstopdf,color,soul}

\usepackage{tabularx}
\usepackage[ruled,vlined]{algorithm2e}
\allowdisplaybreaks

\usepackage{enumerate}
\usepackage[shortlabels]{enumitem}
\usepackage{multicol}
\usepackage{empheq}
\usepackage{caption} 
\usepackage{float}

\usepackage[dvipsnames]{xcolor}
\definecolor{blue}{gray}{0}

\newcommand{\Eset}{\mathbb{E}}

\newcommand{\Rset}{\mathbb{R}}

\newcommand{\Bcal}{{\cal B}}

\newcommand{\Ecal}{{\cal E}}
\newcommand{\Fcal}{{\cal F}}
\newcommand{\Gcal}{{\cal G}}

\newcommand{\Ncal}{{\cal N}}
\newcommand{\Ocal}{{\cal O}}

\newcommand{\Tcal}{{\cal T}}

\newcommand{\Vcal}{{\cal V}}

\newcommand{\Xcal}{{\cal X}}

\newcommand{\Abf}{{\bf A}}

\newcommand{\Ibf}{{\bf I}}

\newcommand{\Wbf}{{\bf W}}

\newcommand{\Ybf}{{\bf Y}}
\newcommand{\Zbf}{{\bf Z}}

\newcommand{\btheta}{{\bar{\theta}}}

\newcommand{\bfTheta}{{\boldsymbol{\Theta}}}

\newcommand{\1}{{\mathbf{1}}}

\newtheorem{lem}{Lemma}
\newtheorem{thm}{Theorem}

\newtheorem{assump}{Assumption}

\newcommand{\vct}[1]{\boldsymbol{#1}}
\newcommand{\mtx}[1]{\boldsymbol{#1}}

\newcommand{\mF}{\mtx{F}}

\newcommand{\mX}{\mtx{X}}

\usepackage{tikz}

\title{Finite-Time Convergence Rates of Decentralized Stochastic Approximation with Applications in Multi-Agent and Multi-Task Learning}
\author{Sihan Zeng\thanks{School of Electrical and Computer Engineering, Georgia Institute of Technology, Atlanta, GA
.}
\and Thinh T. Doan\thanks{Bradley Department of Electrical and Computer Engineering, Virginia Tech, Blacksburg, VA.}
\and Justin Romberg\footnotemark[1]}
\let\underbrace\LaTeXunderbrace

\begin{document}

\maketitle
\begin{abstract}

We study a decentralized variant of stochastic approximation, a data-driven approach for finding the root of an operator under noisy measurements. A network of agents, each with its own operator and data observations, cooperatively find the fixed point of the aggregate operator over a decentralized communication graph. Our main contribution is to provide a finite-time analysis of this decentralized stochastic approximation method when the data observed at each agent are sampled from a Markov process; this lack of independence makes the iterates biased and (potentially) unbounded. 
Under fairly standard assumptions, we show that the convergence rate of the proposed method is essentially the same as if the samples were independent, differing only by a log factor that accounts for the mixing time of the Markov processes. The key idea in our analysis is to introduce a novel Razumikhin-Lyapunov function, motivated by the one used in analyzing the stability of delayed ordinary differential equations. 
We also discuss applications of the proposed method on a number of interesting learning problems in multi-agent systems.
\end{abstract}

\section{INTRODUCTION} 
Stochastic approximation (SA), introduced by \cite{RobbinsM1951}, is a simulation-based approach for finding the root (or fixed point) of some unknown operator under corrupted measurements. This method has broad applications in many areas including stochastic optimization \cite{Nemirovski_SA_2009} and reinforcement learning \cite{SBbook2018}, as existing algorithms in these areas can be viewed as different variants of SA. In this paper, our interest is to study a decentralized stochastic approximation (DCSA) framework, where there are a group of $N$ agents interacting through a decentralized communication network\footnote{We use ``agent" to represent for a sensor, a robot, or a smart device.}. The goal of the agents is to collaboratively find the root of a global operator composed of local operators distributed among the agents. 

Mathematically, each agent $i$ has access to a local operator $F_i:\Xcal_i\times\mathbb{R}^{d}\rightarrow\mathbb{R}^{d}$, where $\Xcal_i$ is a statistical sample space with distribution $\mu_i$. The goal of the agents is to find $\theta^*$ satisfying
\begin{align}
    \bar{F}(\theta^*)\triangleq\sum_{i=1}^{N}\bar{F}_i(\theta^*)=0,
    \label{eq:objective}
\end{align}
where $\bar{F}_i$ is defined as 
\begin{align*}
    \bar{F}_i(\theta)\triangleq \mathbb{E}_{\mu_i}[F_i(X_i, \theta)]=\sum_{X_i\in\Xcal_i}\mu_i(X_i)F_i(X_i,\theta).
\end{align*}
{\color{blue} We assume that at least one solution $\theta^*$ to \eqref{eq:objective} exists, and the structure of $\bar{F}$ that we discuss later implies that $\theta^*$ is unique.}
The agents can interact with each other through an undirected and connected graph $\Gcal=\left(\Vcal,\Ecal\right)$, where agents $i,j\in\Vcal$ communicate if and only if $(i,j)\in\Ecal$. We denote by $\Ncal(i)$ the neighboring set of agent $i$, i.e. $\Ncal(i)\triangleq\{j\mid(i,j)\in\Ecal\}$.\looseness=-1




We are motivated by the wide applications of decentralized framework in many areas due to the rapid development of wireless sensor networks, mobile robots, and the Internet of Things. In these areas, since the data are already collected in a decentralized manner, decentralized approach for solving this problem is preferable to its centralized counterpart. For example, when the data dimension is very large (e.g., for image or video sensors) transmitting the raw data to a fusion center for processing is time consuming and requires massive communications and large bandwidths. 
In addition, mobile sensors usually have limited communication capacity and are only able to communicate with other agents nearby, for example, in the application of using unmanned robots with limited energy to collaboratively explore an unknown terrain or boundary \cite{ovchinnikov2014decentralized}. 

\begin{algorithm}[ht]
\SetAlgoLined
\textbf{Initialization:} Agent $i$ initializes $\theta_i^0\in\Rset^d$ and $\{\epsilon_k\}_{k\in\mathbb{N}}$.

 \For{k=0,1,2,...}{
  \For{each agent $i$}{
    1) Exchange $\theta_i^k$ with its neighbors $j\in\Ncal_{k}(i)$\\
    2) Update the local iterates:
        \begin{align}
        \hspace{-5pt}\theta_{i}^{k+1}=\hspace{-5pt}\sum_{j\in\Ncal(i)}W(i,j)\theta_{j}^{k}+\epsilon_k F_i\left(X_{i}^{k}, \theta_{i}^{k}\right)
        \label{Alg:DCSA:Update}
        \end{align}
  }
 }
\caption{Decentralized Stochastic Approximation}
\label{Alg:DCSA}
\end{algorithm}

Our proposed DCSA for solving \eqref{eq:objective} is formally stated in Algorithm \ref{Alg:DCSA}, where each agent $i$ maintains $\theta_i^k$, a local estimate of the optimal $\theta^*$.  In every iteration $k$, the iterate $\theta_i^k$ is updated according to $\eqref{Alg:DCSA:Update}$, a combination between a consensus step and a local SA step. In the consensus step, agent $i$ collects its neighbors parameters $\{\theta_j^k\mid j\in\Ncal_i\}$ and performs a weighted average using the weights $W(i,j)$ to achieve an agreement between the agents' iterates. On the other hand, to push the consensus point toward $\theta^*$ agent $i$ moves in the direction of its local operator $F_i(X_i^k,\theta_i^k)$ scaled by some step size $\epsilon_k$.

Our focus is to study the finite-time performance of DCSA when the data $\{X_i^k\}$ observed at each agent $i$ are sampled from Markov processes. This Markov sampling results to the data being dependent and biased. The existing theoretical guarantees for the finite-time performance of DCSA are studied under the common assumption that the local data at each agent are independent and identically distributed (i.i.d.). Such an assumption may not hold in many applications, where the data are dependent and evolve over time, for example, when they are sampled from some dynamical systems like in stochastic control or reinforcement learning. Our goal is to close this gap.

\subsection{Main Contribution}
The main contribution of this work is to establish the finite-time convergence rates of DCSA (Algorithm~\ref{Alg:DCSA}) when the data at each agent are generated by Markov processes. We do so under general conditions on the local operators $F_{i}$ (Lipschitz continuity) and the global operator ($1$-point strongly monotone). In particular, we provide an explicit formula that characterizes the convergence rates of Algorithm \ref{Alg:DCSA} for both constant and time-varying step sizes. Our results, explained in details in Section \ref{sec:finite_time_analysis}, differ from the i.i.d.\ case by only a log factor, which naturally arises due to the geometric mixing time of the Markov chain. In addition, our rate matches the one in centralized stochastic approximation with Markovian data up to a factor that depends on the structure of the communication graph $\Gcal$. This paper significantly extends our preliminary work in \cite{zeng2021finite} as discussed in details in Section~\ref{sec:related_work}.

Key to our analysis is a novel Lyapunov function tailored to decentralized algorithms with Markovian samples. Since the data are dependent, the current iterates depend on the history of the algorithm. Thus, the analysis of Algorithm \ref{Alg:DCSA} must carefully account for data across different delays. 
Inspired by the popular use of Razumikhin-Lyapunov functions in studying the stability of delayed ordinary differential equations, we design a Lyapunov function of this type to derive the convergence rates for DCSA. We believe that this Lyapunov approach is new in the literature of decentralized algorithms, and is of independent interest as it may be applied to many problems in this setting.

Finally, we discuss how our theoretical results can be applied to characterize the performance of two important special cases: decentralized Markov chain gradient descent and decentralized Q-learning, and provide some illustrative numerical examples.



\subsection{Related Work}\label{sec:related_work}
In this section, we provide a brief summary of the existing literature in decentralized stochastic gradient descent (DSGD) and decentralized reinforcement learning methods, both of which can be viewed as special cases of DCSA. This will help to distinguish our paper with the existing work in these areas.      


In \cite{lian2017can,zeng2018nonconvex,koloskova2020unified,khaled2020tighter} and the references therein, DSGD is studied for optimizing convex and non-convex problems under i.i.d.\ data (realizations of the gradient).  In \cite{sun2019decentralized,wai2020convergence}, Markovian data are considered, similar to the setting in this paper. The existing theoretical results of DSGD are established based on various assumptions on the boundedness of the gradients/updates. For example, the authors in \cite{sun2019decentralized} assume that the norm of $F_i(X_i,\theta)$ is uniformly bounded, while a bounded variance of $\sum_{i=1}^{N}\left(F_i(X_i,\theta)-\bar{F}_i(\theta)\right)$ is assumed in \cite{wai2020convergence}. Similarly, the bounded variance condition, a bound on the norm of the error between the stochastic gradient and the true gradient, is considered in \cite{lian2017can}. The authors in  \cite{koloskova2020unified} relax this condition to only require that the expected norm of the stochastic gradient is bounded affinely by the norm of the true gradient. However, \cite[Proposition 1]{khaled2020better} shows that none of these conditions in \cite{lian2017can,zeng2018nonconvex,koloskova2020unified,sun2019decentralized,wai2020convergence,khaled2020tighter} holds on a simple objective function. Relaxing these strong conditions is nontrivial, especially in the context of Markovian data and decentralized framework. One of the main contributions of this current paper is to loosen these assumptions considerably; we only assume that $F_{i}$ (which is the gradient map in DSGD) is Lipschitz continuous, which is observed by the example in \cite[Proposition 1]{khaled2020better}.

In the context of reinforcement learning, stochastic approximation is a common tool to analyze the convergence of temporal-difference (TD) learning \cite{srikant2019finite}, Q-learning \cite{chen2019performance}, and other methods \cite{SBbook2018}. In the distributed setting, decentralized TD learning using linear function approximation has been studied in \cite{doan2019finite,doan2021finite}, which can be viewed as a special case of DCSA with $F_i$ being linear operators. Moreover, the authors in \cite{doan2021finite} consider the setting where the linear operators at the agents are the same. Therefore, the techniques in \cite{doan2019finite,doan2021finite} cannot be applied to analyze DCSA, where we consider nonlinear SA and the local operators are different across the agents. Moreover, DCSA captures the decentralized variant of the popular Q-learning {\color{blue} under linear function approximation, whose convergence property is unknown in the existing literature. With an assumption on the state visitation, we show that the problem fits in our analytical framework and its convergence rate can be derived as a consequence of our main theorem. The decentralized policy gradient methods for multi-agent and/or multi-task reinforcement learning in \cite{zhang2018fully,assran2019gossip,zeng2021decentralized,lin2021multi,zeng2021learning,zhang2021finite} also perform updates similar to \eqref{Alg:DCSA:Update} in spirit that combine local gradient steps with network-level information aggregation.}


We also want to point out some related work in the context of parameter/server settings (or federated learning/local SGD) \cite{kairouz2021advances,li2020federated}, where the agents communicate with a centralized coordinator, a problem that has attracted a good deal of interest lately because of its broad applications in data center and cloud computing. This centralized communication setting can be a special case of Algorithm \ref{Alg:DCSA} with a ``star graph''. Recent work on the theoretical convergence of this type of federated learning can be found in \cite{kairouz2021advances,li2020federated,doan2020local} and the references therein. Extending these theoretical results in the centralized communication setting to the decentralized setting here, however, is not straightforward.

Finally, we note that a preliminary version of this work is available in  \cite{zeng2021finite}, where we considered DCSA algorithm under a constant step size\footnote{For convenience, we submit the conference version \cite{zeng2021finite} in the supplementary material of this current paper.}. In the current work, we study both constant and time-varying step sizes and provide a complete analysis for the finite-time convergence rates of DCSA, which is missing in \cite{zeng2021finite}. We also discuss our results when the underlying communication graphs are time-varying, which is not covered in \cite{zeng2021finite}. \looseness=-1


\section{FINITE-TIME ANALYSIS}\label{sec:finite_time_analysis}
In this section, we present the main results of this paper, where we explicitly characterize the convergence rate of Algorithm \ref{Alg:DCSA} in solving \eqref{eq:objective}. 
We provide an upper bound on the error $\Eset\big[\|\theta_{i}^{k}-\theta^*\|^2\big]$ for any $i\in\Vcal$ when Algorithm \ref{Alg:DCSA} uses constant and time-varying step size sequences $\{\epsilon_k\}$. When $\epsilon_{k} = \epsilon$ for some constant $\epsilon$, this error obeys
\begin{align*}
\Eset\big[\|\theta_{i}^{k}-\theta^*\|^2\big]&\leq  \Ocal\left(a^k+\epsilon\log(\frac{1}{\epsilon})\right),
\end{align*}
for a constant $a<1$ explicitly shown later in the section. Under time-varying step sizes $\epsilon_{k}=\Ocal(1/k)$
\begin{align*}
\Eset\big[\|\theta_{i}^{k}-\theta^*\|^2\big]&\leq \Ocal\left(\frac{\log^2(k)}{k}\right).
\end{align*}
We begin by introducing the technical assumptions and useful notation used in our analysis in the following subsections. 
\subsection{Technical Assumptions and Notation}
Given a constant $\epsilon > 0$, we denote by $\tau(i,\epsilon)$ the mixing time associated with the Markov chain $\{X_{i}^{k}\}$, i.e., $\tau(i,\epsilon)$ is the {\color{blue} smallest integer} such that $\forall x_{i}\in\mathcal{X}_{i}$
\begin{align*}
d_{TV}(P(X_i^{k}=\cdot\mid X_i^0=x_i),\mu_i) \leq \epsilon,\quad \forall k\geq \tau(i,\epsilon),
\end{align*}
where  $d_{TV}(\cdot,\cdot)$ is the total variance distance and $P(X_i^{k}=\cdot\mid X_i^0=x_i)$ is the distribution of $X_{i}^{k}$ with the initial conditions $x_{i}$. {\color{blue}The mixing time represents the time it takes for the distribution of $X_{i}^{k}$} to get close to the stationary distribution $\mu_{i}$. We consider the following technical assumption on the properties of the Markov chains $\{X_i^k\}$. 
{\color{blue}
\begin{assump}
\label{assump:X_ergodic}
The space $\Xcal_{i}$ are compact, and the Markov chains $\{X_i^k\}_k$ are ergodic for all $i=1,2,...,N$.
\end{assump}
Assumption \ref{assump:X_ergodic} implies that the Markov chain $\{X_{i}^{k}\}_k$ has a geometric mixing time. In other words, there exist $m_i>0$ and $\rho_i\in(0,1)$ such that $\forall x_{i}\in\mathcal{X}_{i}$
\begin{align}
    d_{TV}(P(X_i^{k}=\cdot\mid X_i^0=x_i),\mu_i)\leq m_i\rho_i^{k},\quad\forall k\geq 0.
    \label{eq:ergodic_consequence}
\end{align}
In the special case where $\Xcal_i$ is finite, the ergodicity of the Markov chain is guaranteed when it is irreducible and aperiodic, and the mixing time depends on the second largest eigenvalue of the transition probability matrix of $\{X_{i}^{k}\}_k$ \cite{LevinPeresWilmer2006}.} 

\eqref{eq:ergodic_consequence} implies that there exists a constant $\beta>0$ such that
\begin{align}
\tau(i,\epsilon)\leq \beta\log(1/\epsilon) \text{ for all } i.
\label{eq:geometric_tau}
\end{align} 
{\color{blue}
Given any $\epsilon>0$ and denoting $\rho\triangleq\max_{i=1,2,...,N}\rho_i$, we define 
\begin{align}
    \tau(\epsilon)\triangleq\max\{\frac{\rho}{1-\rho},\max_{i=1,2,...,N} \tau(i,\epsilon)\},
    \label{eq:tau_def}
\end{align}
For convenience, we denote $\tau_k=\tau(\epsilon_k)$. 
We require $\tau(\epsilon)\geq\frac{\rho}{1-\rho}$ as this condition later helps us to show $\tau_k\leq\tau_{k-1}+1$ for $k>\tau_k$, i.e. $\tau_k$ does not increase too fast.}
In addition, as long as the step size $\epsilon_k$ is a polynomial function of $k$, it is obvious that $\tau_k$ {\color{blue}grows at most logarithmly in $k$}, and we can find a constant $c_{\tau}\in(0,1)$ such that
\begin{align}
    \tau_k+1\leq(1-c_{\tau})(k+1),\quad\forall k>\tau_k.
    \label{eq:k_tauk}
\end{align}
Assumption \ref{assump:X_ergodic} holds in various applications, e.g, in incremental optimization \cite{RamNV2009} and in reinforcement learning modeled by Markov decision processes with a finite number of states and actions.
In addition, Assumption  \ref{assump:X_ergodic} is used in the existing literature to study the finite-time performance of {\sf SA} and its distributed variants under Markov randomness; see  \cite{srikant2019finite,chen2019performance} and the references therein. 

Next, we consider two assumptions on the operators $F_{i}$.
\begin{assump}
\label{assump:F_Lipschitz}
The mappings $F_i(X_i,\cdot)$ are Lipschitz continuous; there exists a constant $L>0$ such that for all $X_i\in\Xcal_i$
\begin{align*}
    \|F_i(X_i,\theta)-F_i(X_i,\tilde{\theta})\|\leq L \|\theta-\tilde{\theta}\|,\quad\forall \theta,\tilde{\theta}\in\mathbb{R}^{d}.
\end{align*}
\end{assump}
\begin{assump}
\label{assump:Fbar_stronglymonotone}
The mapping $-\bar{F}$ is 1-point strongly monotone w.r.t $\theta^*$ i.e. there exists $\alpha>0$ such that
\begin{align}
    \langle\bar{F}(\theta)-\bar{F}(\theta^*),\theta-\theta^*\rangle\leq-\alpha\|\theta-\theta^*\|^2,\quad\forall \theta\in\mathbb{R}^d.\label{assump:Fbar_stronglymonotone:eq}
\end{align}
\end{assump}
Assumption \ref{assump:F_Lipschitz} implies that the operators $F_{i}$ are uniformly Lipschitz continuous, which is weaker than the boundedness assumptions  considered in the existing literature \cite{lian2017can,zeng2018nonconvex,koloskova2020unified}. As a result of Assumption \ref{assump:F_Lipschitz}, the mapping $\bar{F}_i(\cdot)$ is also Lipschitz continuous with constant $L$.
\begin{align*}
    \|\bar{F}_i(\theta)-\bar{F}_i(\tilde{\theta})\|\leq L \|\theta-\tilde{\theta}\|,\quad\forall \theta,\tilde{\theta}\in\mathbb{R}^{d}.
\end{align*}
Since $\Xcal_{i}$ are compact we denote by
\begin{align}
B \triangleq \max\{L, \max_{i,X_i\in\Xcal_i}\|F_i(X_i,0)\|\} < \infty.\label{eq:def_B}
\end{align}
{\color{blue}Under Assumption \ref{assump:Fbar_stronglymonotone}, $\bar{F}_{i}$ is 1-point strongly monotone, which implies that the solution $\theta^*$ of \eqref{eq:objective} is unique. 
This assumption obviously holds in the case where the operator $\bar{F}$ is globally strongly monotone, i.e. $\theta^*$ in \eqref{assump:Fbar_stronglymonotone:eq} can be replaced by any $\theta'\in\mathbb{R}^d$.
The global strong monotonicity is often assumed in the existing literature in studying the finite-time convergence rate of SA; see for example \cite{chen2019performance} and the references therein. In linear SA where the operator $\bar{F}(\theta)=\bar{A}\theta$ for some matrix $\bar{A}\in\mathbb{R}^{d\times d}$, $\bar{A}$ is usually assumed to be negative definite, which implies the strong monotonicity of $\bar{F}$ \cite{srikant2019finite,doan2021finite}.}
In the context of SGD, where $\bar{F}$ is a gradient mapping, this assumption is equivalent to the the restricted Secant inequality \cite{karimi2016linear}, which is a generalized variant of strong convexity and is often used to guarantee the linear convergence of SGD for non-convex optimization.

Finally, we consider the following assumption of the weight matrix of the network $\Wbf$, which encodes the communication structures shared between agents.    
\begin{assump}
\label{assump:W_doublystochastic}
The matrix of consensus weights $\Wbf = [W(i,j)]\in\Rset^{N\times N}$ in \eqref{Alg:DCSA:Update} is doubly stochastic, i.e., $\sum_{i}W_{ij} = \sum_{j}W_{ij} = 1$. 
In addition, $W(i,i) > 0 $ and  $W(i,j)>0$ if and only if $(i,j)\in\Ecal$. 
Otherwise $W(i,j) = 0$. 
\end{assump}
This assumption is a standard condition in the existing literature of decentralized consensus-based algorithms to guarantee that the agents reach an agreement \cite{nedic2018network}. When the underlying network is connected, it implies that the largest singular value of $\Wbf$ is $1$, while other singular values are strictly less than $1$. The rates of  Algorithm \ref{Alg:DCSA} depend on the spectral gap $1-\sigma_2$ representing the connectivity of the network, where $\sigma_2\in[0,1)$ denotes the second largest singular value of $\Wbf$. In the rest of this paper, we assume that Assumptions \ref{assump:X_ergodic}--\ref{assump:W_doublystochastic} always hold. \\ 

\noindent\textbf{Notation}. We use $\|\cdot\|_F$ and $\|\cdot\|_*$ to denote the matrix Frobenius norm and the spectral norm, respectively. We denote by $\1$ and $\Ibf$ the all-one vector in $\Rset^{N}$ and the identity matrix in $\Rset^{N\times N}$. 
Given a sequence of local vectors $\theta_{1},\ldots,\theta_{N}\in\Rset^{d}$, we denote by $\bfTheta$ a matrix in $\Rset^{N\times d}$, whose $i$-th row is $\theta_i^T$,
\begin{align*}
\bfTheta =\left[\begin{array}{cc}
-\;\theta_1^T\;-  \\
\cdots\\
-\;\theta_N^T\;-
\end{array} \right]\in\Rset^{N\times d}.
\end{align*}
We then denote their average by $\btheta$  
\[
\bar{\theta}\triangleq\frac{1}{N}\sum_{i=1}^{N}\theta_i=\frac{1}{N}\bfTheta^{\top}\1\in\mathbb{R}^d.
\]



\subsection{Main Results}
We now present the main result of this paper, where we provide an explicit formula to characterize the finite-time convergence rates of Algorithm \ref{Alg:DCSA} under both constant and time-varying step sizes. {\color{blue}Recalling the definitions of $L$, $\alpha$, $c_{\tau}$, and $B$ from Assumptions~\ref{assump:F_Lipschitz} and \ref{assump:Fbar_stronglymonotone} and Eqs.~\eqref{eq:k_tauk} and \eqref{eq:def_B}, we introduce the following constants for convenience.}
\begin{align}
\begin{aligned}
    &C_1\hspace{-2pt}=\hspace{-2pt}\left(60B^2+\frac{45}{2}+90BL+6B^2\right)\left(\|\theta^*\|^2+1\right),\\
    &C_2\hspace{-2pt}=\hspace{-2pt}\left(\frac{21B}{2}+\frac{5}{6}+\frac{8L^2}{\alpha}+10L\right),\\
    &C_{\epsilon,1}\hspace{-2pt}=\hspace{-2pt}\max\{6B,\frac{45B+132B^2+192BL}{\alpha}\},\\
    &C_{\epsilon,2}\hspace{-2pt}=\hspace{-2pt}\max\{16B,\frac{768B^2}{c_{\tau}\alpha},\frac{\alpha}{4}\hspace{-2pt}+\hspace{-2pt}\frac{128B^2}{c_{\tau}(1\hspace{-2pt}-\hspace{-2pt}\sigma_2^2)}\hspace{-2pt}+\hspace{-2pt}2C_2,\frac{32B^2}{C_2}\}.
    \end{aligned}\label{notation:constant_C}
\end{align}
{\color{blue}Later in Section~\ref{sec:analysis:proof_sketch} we will introduce the notion of optimality error and consensus error. Loosely speaking, $C_2$ describes the coupling between the two errors and $C_1$ represents the effect of the Markov randomness on the optimality error.}

We consider the following two step size sequences.\\
\textbf{Constant Step Size:} Let $\epsilon_k=\epsilon$ ($\tau_k=\tau(\epsilon)$) satisfy
\begin{align}
    \epsilon\tau(\epsilon) &\leq \min\left\{\frac{1}{NC_{\epsilon,1}},~\frac{1-\sigma_2^2}{NC_{\epsilon,2}}\right\}.
    \label{thm:stepsize:constant}
\end{align}
\textbf{Diminishing Step Size:} Let $\epsilon_{k}$ be chosen as 
\begin{align}
    &\epsilon_k=\frac{\epsilon}{k+1},\quad \text{where }\epsilon\geq\frac{8}{\alpha},\notag\\
    &\hspace{-30pt}\text{and }\epsilon_{k-\tau_k}\tau_k\leq\min\left\{\frac{1}{NC_{\epsilon,1}},\frac{1-\sigma_2^2}{NC_{\epsilon,2}}\right\},\quad\forall k\geq\tau_k.\label{thm:stepsize:tv1}
\end{align}

By \eqref{eq:geometric_tau} we have $\lim_{\epsilon\rightarrow 0} \epsilon\tau(\epsilon) = 0$, so one can choose an $\epsilon$ small enough that \eqref{thm:stepsize:constant} is satisfied. A similar argument holds for \eqref{thm:stepsize:tv1}.

\begin{thm}
\label{thm:constant_decay_epsilon_fixedG}
Let $\{\theta_i^k\}$, $\forall i\in\Vcal$, be generated by Algorithm \ref{Alg:DCSA}.\vspace{0.2cm}\\
1. Under the constant step size in \eqref{thm:stepsize:constant}, we have $\forall k>\tau(\epsilon)$
    \begin{align}
        \mathbb{E}\left[\|\theta_i^k-\theta^*\|^2\right]
        &\leq(1-\frac{\alpha\epsilon}{8})^{k-\tau(\epsilon)}\frac{(4S^0+7\|\btheta^0\|^2+C_1}{(1-\sigma_2^2)}+\frac{208N\beta C_1}{\alpha(1-\sigma_2^2)}\epsilon\log(\frac{1}{\epsilon}).\label{thm:rate:constant_stepsize}
    \end{align}
2. Under the diminishing step size in \eqref{thm:stepsize:tv1}, we have $\forall k>\tau_{k}$
    \begin{align}
        \mathbb{E}\left[\|\theta_i^k-\theta^*\|^2\right]&\leq \frac{4(\beta(\log(k+1)-\log(\epsilon))+1)}{(1-\sigma_2^2)(k+1)}\Big(4S^0+7\|\btheta^0\|^2+C_1\Big)\notag\\
        &\hspace{20pt}+\frac{208N\beta C_1\log(k)(\log(k)-\log(\epsilon))}{c_{\tau}^2 (1-\sigma_2^2)(k+1)},\label{thm:rate:tv}
    \end{align}
    where $c_{\tau}\in(0,1)$ is defined in \eqref{eq:k_tauk}, {\color{blue} and $S^0$ is a constant defined later in \eqref{eq:def_S} that depends on the initial condition}.
\end{thm}

With its proof presented in Section~\ref{sec:proof_main}, Theorem \ref{thm:constant_decay_epsilon_fixedG} states that under a constant step size $\epsilon$ every local iterate generated by Algorithm~\ref{Alg:DCSA} converges linearly to a ball around the optimal solution with radius on the order of $\epsilon\log(\frac{1}{\epsilon})$. Under a proper decaying step size, every local iterate converges exactly to the optimal solution with rate $\Ocal(\log^2(k)/k)$.

When $F_{i}$ is the gradient of some function, our result resembles the best known convergence rate of decentralized {\sf SGD}  for solving strongly convex objective functions under i.i.d. samples, differing only by a log factor (for example, see {\color{blue}\cite{sayin2017stochastic}[Theorem 2]}). In addition, this rate matches the ones in \cite{doan2020local} where the authors consider a centralized communication topology. \looseness=-1

Our bound depends inversely on $1-\sigma_2^2$, which reflects the impact of the graph $\Gcal$ and the matrix $\Wbf$ on the convergence. The bound also scales proportional to $N$, which shows the increasing difficulty of the problem as the number of agents goes up.\looseness=-1

\subsection{Main Technical Challenges}
\label{sec:analysis:proof_sketch}
One may attempt to apply the existing analysis in both decentralized optimization and reinforcement learning literature  \cite{nedic2018network,doan2021finite} to derive the convergence rates presented in Theorem \ref{thm:constant_decay_epsilon_fixedG}. However, due to the general settings we consider, in particular, the interaction between the decentralized framework and nonlinear stochastic approximation under Markovian data and (potentially) unbounded operators, such an extension is nontrivial. In the following, we present the main technical challenges in our analysis and explain in detail the reason why the existing technique fails. We then discuss the novel idea in our analysis, where we introduce a variant of Razumikhin-Lyapunov functions \cite{HLbook}. Our complete analysis is given in Section \ref{sec:proof_main}. \looseness=-1

%
%

Our goal is to study the convergence of every iterate $\theta_{i}^{k}$ to the fixed point $\theta^*$. Due to the consensus update, one can separate the error $\theta_{i}^{k} - \theta^*$ as  
\begin{align*}
\theta_{i}^{k} - \theta^* = \underbrace{\theta_{i}^{k} - \btheta^{k}}_{\text{``consensus error"}} +  \underbrace{\btheta^{k} - \theta^*}_{\text{``optimality error"}},
\end{align*}
and study the convergence of these two errors to zero separately, as often done in the decentralized optimization literature \cite{nedic2018network}. However, there are two main challenges that make this approach inapplicable to our setting. For notation simplicity, we define the squared norm of the optimality error in iteration $k$ as
\begin{align*}
    R^{k} \triangleq \|\btheta^{k}-\theta^*\|^2
\end{align*}
and the variance of consensus error in iteration $k$ as
\begin{align}
    S^k \triangleq \sum_{i=1}^N\|\theta_i^k-\btheta^{k}\|^2=\|\bfTheta^k-\1(\btheta^k)^T\|_F^2.\label{eq:def_S}
\end{align}
\textbf{Unbounded Operators}: 
If the operators $F_{i}$ were bounded, i.e., there existed a constant $D$ such that $\|F_i(X,\theta)\|\leq D$ for all $i$, $X$, and $\theta$, one can show that the norm of the consensus error is proportional to the step size $\epsilon$, $S^{k} \sim D\epsilon^2$, which decays to zero as $\epsilon\rightarrow 0$ \cite{nedic2018network}. Based on the convergence of $S^{k}$ the recursion on $R^k$ can be easily established. Here we make the less restrictive assumption that $F_{i}$ is Lipschitz continuous (Assumption \ref{assump:F_Lipschitz}), allowing $\|F_i(X,\theta)\|$ to be unbounded in general. 
In this case, one cannot immediately show that the consensus error converges to zero. Indeed, this error is proportional to the optimality error\looseness=-1
\[
S^{k} \lesssim \delta S^{k-1} + \mathcal{O}(\epsilon^2)R^{k-1}\quad a.s., \quad \text{for some } \delta < 1,
\]
which can diverge unless $R^{k}$ decays to zero at a proper rate.\\ 
\textbf{Markovian Samples}: The data in our stochastic model are sampled from Markov processes. In other words, $\{X_i^k\}_k$ are dependent and have a shift in distribution from sample to sample. As a consequence, $F_{i}(X_{i},\theta_{i})$ is biased, i.e., $\Eset_{\mu_i}[F_{i}(X_{i},\theta_{i})]\neq \bar{F}_{i}(\theta_{i})$, and automatic variance reduction does not follow by simply averaging across consecutive samples. In order to treat the Markovian data, one can utilize the recent development in analyzing (decentralized) stochastic approximation under Markov samples (e.g., \cite{srikant2019finite,doan2021finite}). Specifically, one can take advantage of the geometric mixing time of the underlying Markov chain to handle the data dependence. Although the data are Markovian, their dependence is very weak at samples spaced out every $\tau_k$ step. However, this technique causes an extra ``delay" consensus error in our analysis, i.e., 
\begin{align*}
\Eset[R^{k}] \lesssim \big(1-\frac{\alpha\epsilon}{2}\big)R^{k-1} + \mathcal{O}(\epsilon)\Eset[S^{k-1}] + \Ocal(\epsilon)\Eset[S^{k-\tau_k}].   
\end{align*}
The delay term, $S^{k-\tau_k}$, implies the errors in the earlier iterates affect the current updates. This together with the coupling between consensus and optimality errors above complicates the analysis of Theorem \ref{thm:constant_decay_epsilon_fixedG}. We note that the prior work \cite{doan2021finite} does not face these technical challenges since they consider a special linear case, i.e., $F_{i}$ are linear and the same for every agent. In that case, the consensus error is only proportional to $\epsilon$, similar to the existing literature \cite{nedic2018network}. In addition, there does not exist the delay term $S^{k-\tau_k}$ in the analysis in \cite{doan2021finite}. Without these coupling and delay terms, one can immediately see that the convergence of the consensus error implies the convergence of the optimality error to zero.\\        
\textbf{Novel Lyapunov Approach}: As mentioned above, one cannot study the convergence of consensus and optimality errors separately since these errors are coupled with each other. In addition, one needs to consider the impact of delays due to the Markovian samples. To handle this mixture of errors, a natural way is to analyze these errors simultaneously. Indeed, one of the main contributions in this paper is to introduce a novel \textit{composite Razumikhin-Lyapunov function}  $V$ defined as 
\begin{align*}
    V^k = \mathbb{E}\left[R^{k}\right] + \mathbb{E}\left[S^k\right] + \mathbb{E}\left[S^{k-\tau_k}\right].
\end{align*}
Our main motivation in using $V^{k}$ comes from the wide application of Razumikhin-Lyapunov function in studying the stability of delayed ordinary differential equations \cite{HLbook}. By putting together the bounds of the optimality and consensus errors mentioned above, we will show that $V^{k}$ decays to zero at a proper rate, which immediately implies the results in Theorem \ref{thm:constant_decay_epsilon_fixedG}. 

\subsection{Time-Varying \& Directed Communication Graphs}\label{sec:TV_Graphs}
We note that with slight modifications, our algorithm and theoretical results extend to the case where the agents communicate according to a time-varying communication graph $\Gcal_k=(\Vcal,\Ecal_k)$. Essentially, we assume that the there exists a positive integer $\Bcal$ such that for any integer $l\geq 0$, the following graph is connected
\begin{equation*}
    \left(\Vcal,\Ecal_{l\Bcal}\cup\Ecal_{l\Bcal+1}\cup...\cup\Ecal_{(l+1)\Bcal-1}\right).
\end{equation*}

This assumption is often considered in prior works that study consensus algorithms under a time-varying graph \cite{nedic2018network, doan2019finite}, and guarantees the long-term connectivity though the network in any step can be disconnected. Let the consensus weight matrix $\Wbf_k$ be any double stochastic matrix that respects the graph $\Gcal_k$ (see Assumption \ref{assump:W_doublystochastic}). Our update rule in \eqref{Alg:DCSA:Update} becomes
\begin{align}
    \theta_{i}^{k+1}=\sum_{j\in\Ncal_{k}(i)}W_{k}(i,j)\theta_{j}^{k}+\epsilon_k F_i\left(X_{i}^{k}, \theta_{i}^{k}\right),
    \label{Alg:DCSA:Update:Timevarying}
\end{align}
where $\Ncal_k(i)$ denotes the neighbors of agent $i$ in iteration $k$.

It has been observed in \cite{nedic2018network} that under a time-varying communication graph, the key structural quantity affecting the rate of convergence is 
\begin{equation*}
    \eta=\min\{1-1/(2N^3), \min_{\ell\in\mathbb{Z}_+}\max_{(\ell-1)\Bcal \leq t'\leq \ell\Bcal-1}\hspace{-5pt}\sigma_2(\Wbf_{t'})\}\in[0,1).
\end{equation*}
$\eta$ plays the role of $\sigma_2$ in the fixed communication graph setting. Our analysis can be adapted to show that with a time-varying communication graph, the modified update rule \eqref{Alg:DCSA:Update:Timevarying} produces iterates that converge linearly to a ball around the optimal solution $\theta^*$ under a constant step size; under a diminishing step size $\epsilon_k=\Ocal(1/k)$, the iterates converge exactly to $\theta^*$ with rate $\Ocal(\log^2(k)/k)$. These rates are the same as our results under a fixed communication graph, except for the constant $\Bcal$ capturing the connectivity of time-varying graphs.

Similarly, our analysis can be extended to the case where the communication graph is directed and strongly connected by leveraging the techniques in \cite{nedic2014distributed}.

\section{APPLICATIONS \& EXPERIMENTS}
\label{sec:applications}
In this section, we illustrate our theoretical results by considering a number of numerical simulations on the performance of decentralized Markov chain gradient descent and decentralized Q-learning methods in solving problems in robust identification and reinforcement learning. We show that these two methods can be formulated as different variants of Algorithm \ref{Alg:DCSA}.

\subsection{Decentralized Markov Chain Gradient Descent}
We consider a distributed optimization problem over a network of $N$ agents. Associated with agent $i$ is a function $\bar{f}_i:\Rset^{d}\rightarrow \Rset$. The goal of the agents is to cooperatively solve
\begin{align}
    \min_{\theta \in \mathbb{R}^d} \sum_{i=1}^{N} \bar{f}_i(\theta),
    \label{eq:decentralized_MCGD_objective}
\end{align}
where $\bar{f}_i(\theta) = \mathbb{E}_{X_i\sim\mu_i}\left[f_i(X_i, \theta)\right]$ for some $f_i:\Xcal_i\times \Rset^{d}\rightarrow \Rset$ and $X_i$ is a random variable taking values over the space $\Xcal_i$ with the stationary distribution $\mu_i$. Decentralized stochastic gradient descent is one of the most popular methods in the literature for solving this problem, where each agent $i$ maintains $\theta_{i}$, an estimate of the solution $\theta^*$ of \eqref{eq:decentralized_MCGD_objective}, and iteratively performs\looseness=-1
\begin{align}
\theta_{i}^{k+1} = \sum_{j\in\Ncal_{i}}W_{ij}\theta_{j}^{k} + \epsilon_{k}\nabla f_{i}(X_{i}^{k},\theta_{i}^{k}). \label{eq:DCSGD}    
\end{align}
Most of the existing results about the convergence of \eqref{eq:DCSGD} are established under the conditions that $X_{i}^{k}$ are i.i.d. across time and $i$. However, in many problems such as optimization under a linear dynamic system, $\{X_i^{k}\}_k$ are sampled from Markov chains. See \cite{sun2018markov,sun2019decentralized} for a discussion on more applications where the data are naturally Markovian.

\begin{figure}[h]
\centering  \includegraphics[width=.9\linewidth]{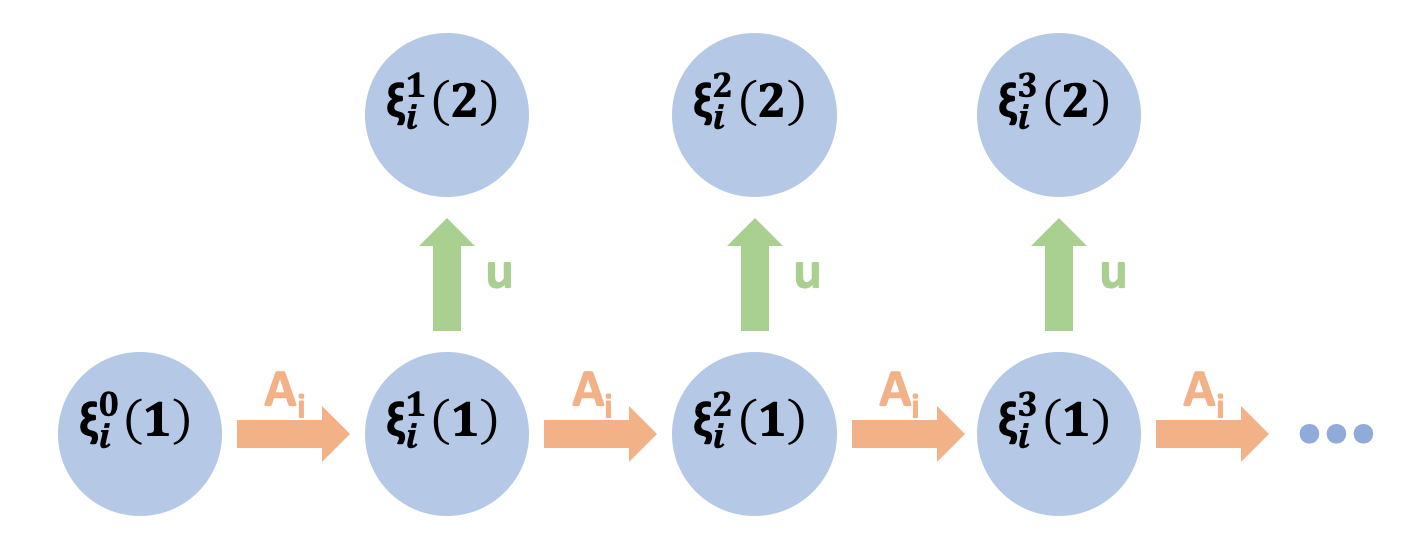}
  \caption{System ID Sample Generation}
  \label{figure:SystemID_sample}
  \vspace{-.3cm}
\end{figure}

As a concrete example, we consider the distributed variant of the robust identification problem \cite{POLJAK198053}. In this problem, each agent $i$ is associated with a system as shown in Figure~\ref{figure:SystemID_sample}. The system parameters are $\Abf_i\in\mathbb{R}^{d\times d}$, which is different across agents, and $u\in\mathbb{R}^d$, the same across agents. The observations (samples) are $X_{i}^{k} = (X_{i}^{k}(1),X_{i}^{k}(2))\in\Rset^{d}\times\Rset$, generated by the following auto-regressive process
\begin{align}
\begin{aligned}
&X_{i}^{k}(1) = \Abf_{i}X_{i}^{k-1}(1) + U_{i}^{k},\\
&X_{i}^{k}(2) = \langle u ,X_{i}^{k}(1)\rangle + y_{i}^{k},
\end{aligned}\label{sec:simulation:xi12}
\end{align}
where $U_{i}^{k}\in\mathbb{R}^d$ and $y_{i}^{k}\in\mathbb{R}$ model i.i.d. errors. Obviously, since $\{U_{i}^{k}\}$ and $\{y_{i}^{k}\}$ are i.i.d. $\{X_{i}^{k}\}$ is a Markov chain. Our aim is to estimate the system parameters from the Markovian samples. This problem is considered in \cite{kim2012autonomous}, where the system parameters encode the stability of a civil structure and the authors seek to estimate the parameters using a decentralized network of sensors. Besides civil structure monitoring, this problem also abstracts many other problems in various engineering fields, including epidemic modeling and resource allocation \cite{ige2020markov,zhou2018markov}. 


For simplicity, in this work we only aim to estimate the parameter $u$, which amounts to solving the optimization problem \looseness=-1
\begin{align}
\underset{\theta\in\Rset^{d}}{\text{minimize }} \sum_{i=1}^{N}\bar{f}_{i}(\theta) = \sum_{i=1}^{N} \Eset_{X_{i}\sim\mu_i} \left[f_{i}(X_{i},\theta)\right],\label{sec:simulation:obj}    
\end{align}
where $f_{i}(X_{i},\theta)$ is the quadratic lost function
\begin{align*}
    f_{i}(X_{i},\theta) = \big(\langle \theta,X_{i}(1)\rangle - X_{i}(2)\big)^2.
\end{align*}
We note that the estimate of $u$ is more robust and accurate by using samples across all agents.

To solve problem \eqref{sec:simulation:obj} we use the decentralized Markov SGD \eqref{eq:DCSGD}, a variant of Algorithm \ref{Alg:DCSA}. In this case, $-\bar{F}_{i} = \nabla_{\theta}f_{i}$ is strongly monotone if $\Eset_{X_{i}\sim\mu_i} \left[X_i(1)X_i(1)^{\top}\right]$ is positive definite. Under the strong monotonicity of $-\bar{F}_i$, our theoretical results in Theorem \ref{thm:constant_decay_epsilon_fixedG} imply the same results for the finite-time performance of the decentralized Markov SGD. Our rates recover the best known result of decentralized Markov SGD while relying on weaker assumptions \cite{sun2019decentralized}.


To illustrate our theoretical results, we experimentally solve this problem where we choose $N=50$ and $d=20$. To generate data, we choose $\Abf_{i}\in\Rset^{d\times d}$ to be a subdiagonal matrix where for all $2\leq m\leq d$, $\Abf_{i}[m,m-1]$ is drawn uniformly from $[.8,.99]$, and $u$ to be a random vector drawn uniformly from the d-dimensional unit $\ell_{2}$-ball. We sample scalars $u_i^k$ and $y_i^k$ i.i.d. from the standard normal distribution $N(0,1)$, and make $U_i^k=u_i^k e_1$, where $e_1$ is the first basis vector. We then generate the samples by making $X_i^0(1)=0$ and following \eqref{sec:simulation:xi12} for $k\geq 1$. 
In this experiment, we use a line graph $\Gcal$, i.e. agent $i$ connects to agent $i-1$ and $i+1$ for all $i=2,3,...,N-1$, and agents $1$ and $N$ only connect to agents $2$ and $N-1$, respectively. In addition, we generate $\Wbf$ using the lazy Metropolis method \cite{olshevsky2015linear}.

\begin{figure}[h]
  \centering
  \includegraphics[width=.5\linewidth]{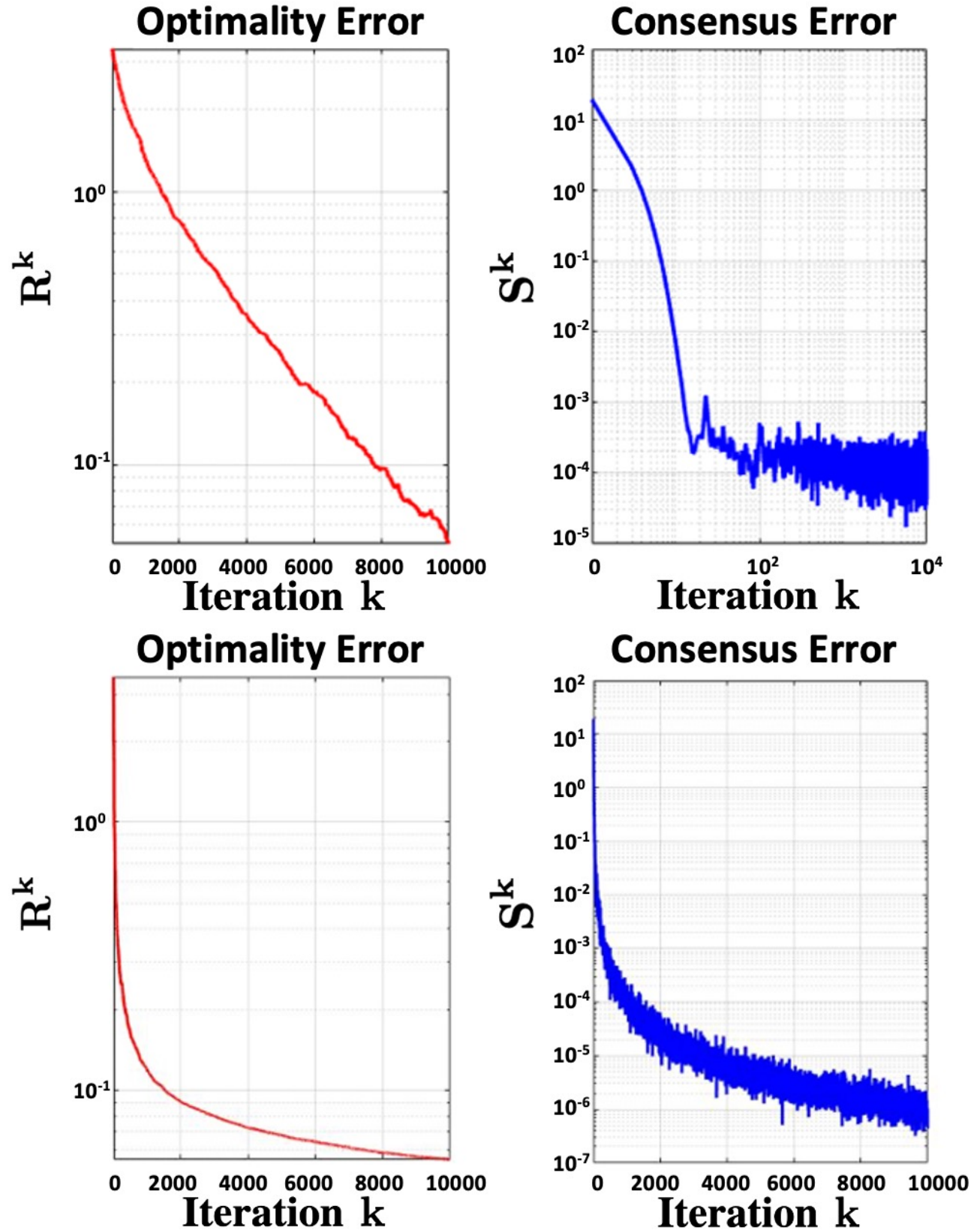}
  \caption{Performance of Markov Chain Gradient Descent on Distributed System Identification. Top Row: Constant Step Size. Bottom Row: Diminishing Step Size}
  \label{figure:SystemID}
  \vspace{-.4cm}
\end{figure}

In Figure \ref{figure:SystemID}, we plot the decay of the errors under a constant step size $\epsilon=5e-4$ (top row) and a diminishing step size with $\epsilon=3e-2$ (bottom row). In the case of the constant step size, we see that the optimality error roughly converges linearly after the initial 2000 iterations, which matches the theory. The decay of the consensus error is much faster, which is a common result and observation in the literature of consensus algorithms \cite{nedic2018network}. In the case of diminishing step size, both the optimality error and the consensus error decays roughly with rate $\Ocal(\log^2(k)/k)$, again matching our theoretical result.\looseness=-1

\subsection{Decentralized Q Learning}
\label{sec:mtql}
We consider multi-task reinforcement learning problems over multi-agent systems, where there are $N$ different tasks and $N$ agents \cite{zeng2021decentralized}. Each task, modeled by a Markov decision process, is assigned to one agent. The goal of the agents is to cooperatively find a single policy, a mapping from state spaces to the action spaces, that simultaneouly solves all the tasks. \looseness=-1

For solving this problem, we consider a decentralized variant of Q-learning method under linear function approximation \cite{chen2019performance}. When the tasks share the same dynamic, this problem is equivalent to find a $\theta^*$ that solves
\begin{align*}
    \sum_{i=1}^{N}\mathbb{E}_{X\sim\mu_i}G_{i}(\theta^*,X)=\vct{0}
\end{align*}
for some non-linear operator $G_i$ that depends on the basis of the linear subspace and the reward functions at different tasks. For more details, we refer the readers to \cite{chen2019performance} which discusses the problem in the single agent case. {\color{blue} Under a proper assumption on $\mu_i$ and $G_i$ that guarantees sufficient state visitation, one can show that the operator $-\bar{G}_i(\theta)=-\mathbb{E}_{X\sim\mu_i}[G_{i}(\theta,X)]$ is strongly monotone. With $F_{i} = -G_{i}$,} our results in Theorem \ref{thm:constant_decay_epsilon_fixedG} can be applied to derive the rates of decentralized Q-learning, which is unknown in the existing literature.

\begin{figure}[h]
  \centering
  \includegraphics[width=.95\linewidth]{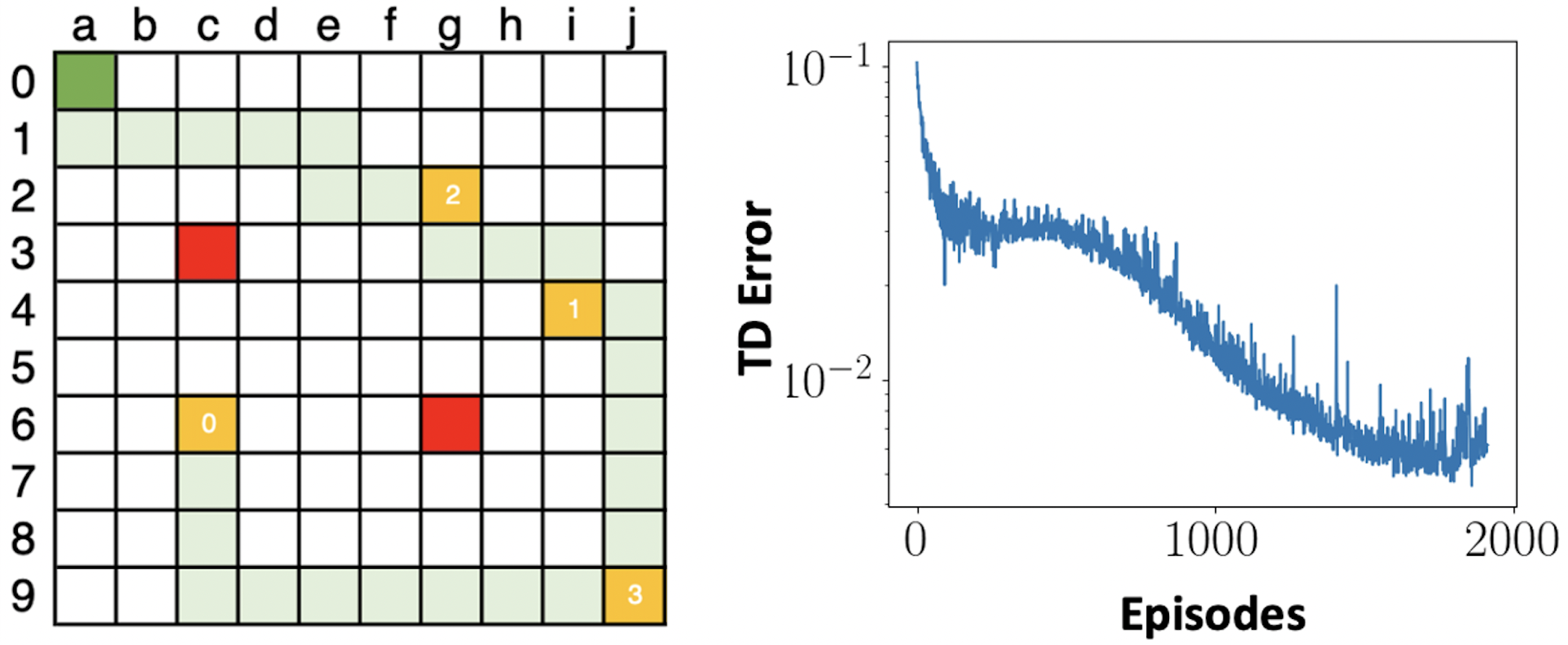}
  \caption{Multi-task Maze Navigation. Left: Path Taken by Learned Policy. Right: Convergence of TD Error}
  \label{figure:DCQL_maze}
  \vspace{-.4cm}
\end{figure}

To illustrate for the performance of decentralized Q-learning, we consider a GridWorld problem consisting of multiple mazes (environments) with size $10\times 10$ (see the first plot of Fig. \ref{figure:DCQL_maze}). 
In each environment, the agent is placed in a grid of cells, where every cell has one of three labels: goal, obstacle, or empty. The agent selects an action from a set of 4 actions {up, down, left, right} to move to the next cell. It then receives a reward of +1 if it reaches the goal, -1 if it encounters an obstacle, and 0 otherwise. The state is the current position of the agent. Each agent maximizes its personal cumulative award when the goal is reached from the initial position in the smallest number of steps. The goal of the agents is to find a single policy that can be applied to all the maze environments.


We connect the agents in a line graph and train them with the decentralized Q-learning for 2000 episodes. At the end of the training phase, the agents agree on a unified policy, whose performance is tested in all environments. The results are presented in the first plot of Figure \ref{figure:DCQL_maze}, where we combine all the results into one grid. Here the dark green cells represent the starting position of the agents. Yellow cells represent the goals, with the white numbers indicating the indices of environments in which the cell is a goal. Red cells represent the obstacle positions. The light green path is the route generated by the learned policy returned by the decentralized Q-learning.


Additionally, we visualize the error decay for this problem under time-varying step sizes in the second plot of Figure~\ref{figure:DCQL_maze}. Since the optimal $\theta^*$ is unknown, we use the temporal-difference error, which is often used as a surrogate convergence metric in reinforcement learning \cite{SBbook2018}. As the TD error converges to zero, the iterates converge to the optimal $\theta^*$. Here, we see the decay of the TD error, though noisy, roughly decays at a rate $\Ocal(1/k)$ after the initial 500 episodes.

\section{Proof of Main Theorem}
\label{sec:proof_main}

\subsection{Preliminaries}
\label{sec:proof_main:preliminaries}

We use the following notations in the rest of the paper.
\begin{align*}
\mX &\triangleq \left[X_1,X_2,...,X_N\right],\\
\mF(\mX;\bfTheta) &\triangleq\left[\begin{array}{cc}
-\;F_1(X_1,\theta_1)^{\top}\;-  \\
\cdots\\
-\;F_N(X_N,\theta_N)^{\top}\;-
\end{array} \right]\in\Rset^{N\times d}.
\end{align*}


The following technical lemmas will be applied regularly in the later analysis. Lemmas \ref{lem:theta_k+theta_k-tau}-\ref{lem:optimalityerror} are established under either the constant step size \eqref{thm:stepsize:constant} or the diminishing step size \eqref{thm:stepsize:tv1}.

\begin{lem}
\label{lem:B_bounded}
Recall $B = \max\{L, \max_{i,X_i\in\Xcal_i}\|F_i(X_i,0)\|\}$. Under Assumption \ref{assump:F_Lipschitz}, we have for all $i$ and $X_i$
\begin{align*}
    &\|F_i(X_i,\theta)\|\leq B\left(\|\theta\|+1\right),\\
    \text{and}\quad &\|\bar{F}_i(\theta)\|\leq B\left(\|\theta\|+1\right),\quad\forall\theta\in\mathbb{R}^{d}.
\end{align*}
\end{lem}
Lemma \ref{lem:B_bounded} is a simple consequence of the Lipschitz condition of $F_i(X_i,\cdot)$, and states that the norm of the operator $F_i(X_i,\theta)$ or $\bar{F}_i(\theta_i)$ can be bounded affinely in $\theta$.

\begin{lem}\label{lem:theta_k+theta_k-tau}
a) For all $k\geq\tau_k$, we have
\begin{align}
    \|\btheta^k-\btheta^{k-\tau_k}\| &\leq 3\epsilon_{k-\tau_k} B\tau_k\left(\|\btheta^k\|+\frac{1}{N}\sqrt{S^{k-\tau_k}}+1\right)\notag\\
    &\leq \frac{1}{2N}\left(\|\btheta^k\|+\sqrt{S^{k-\tau_k}}+1\right).
    \label{lem:theta_k+theta_k-tau:res1}
\end{align}

b) For all $k\leq\tau_k$, we have
\begin{align}
    \|\btheta^k-\btheta^{0}\|\leq \frac{1}{3N}\|\btheta^{0}\|+\frac{1}{3N}\sqrt{S^{0}}+\frac{1}{3N}.
    \label{lem:theta_k+theta_k-tau:res3}
\end{align}
\end{lem}

Lemma \ref{lem:theta_k+theta_k-tau} bounds the distance between the average of the iterates within $\tau_k$ steps.

\begin{lem}\label{lem:consensuserror}
We have for all $k\geq 0$ 
\begin{align*}
S^k&\leq \frac{1+\sigma_2^2}{2}S^{k-1}+\frac{32\epsilon_{k-1}^2 B^2 N}{1-\sigma_2^2}R^{k-1}+\frac{NC_0}{1-\sigma_2^2}\epsilon_{k-1}^2,
\end{align*}
where the constant
\begin{align}
    C_0=16B^2\left(\|\theta^*\|^2+1\right).\label{eq:def_C0}
\end{align}
\end{lem}

\begin{lem}\label{lem:optimalityerror}
We have for all $k\geq \tau_k$
\begin{align*}
    \mathbb{E}[R^{k+1}] &\leq(1-\frac{\alpha\epsilon_k}{2})\mathbb{E}[R^k]+NC_1 \epsilon_k\epsilon_{k-\tau_k}\tau_k+NC_2 \epsilon_k\mathbb{E}[S^k]+NC_2 \epsilon_k\mathbb{E}[S^{k-\tau_k}],
\end{align*}
where $C_1$ and $C_2$ are defined in \eqref{notation:constant_C}.
\end{lem}

Lemmas \ref{lem:consensuserror} and \ref{lem:optimalityerror} characterize the decay of the consensus error and the optimality error, respectively. For conciseness, the proofs of the lemmas are deferred to the appendix.

\subsection{Proof of Theorem \ref{thm:constant_decay_epsilon_fixedG}}
\label{sec:proof_main:theorem}

$V^k = \mathbb{E}[R^{k}]+\mathbb{E}[S^{k}]+\mathbb{E}[S^{k-\tau_{k}}]$ is the sum of three components. Bounding $V^k$ requires characterizing the growth of each of the three terms, which are all coupled with each other.

By Lemmas \ref{lem:consensuserror} and \ref{lem:optimalityerror}, we can recursively bound the consensus error
\begin{align}
    \mathbb{E}[S^k]&\leq \frac{1+\sigma_2^2}{2}\mathbb{E}[S^{k-1}]+\frac{32\epsilon_{k-1}^2 B^2 N}{1-\sigma_2^2}\mathbb{E}[R^{k-1}]+\frac{NC_0}{1-\sigma_2^2}\epsilon_{k-1}^2,
    \label{thm:decayepsilon:ineq0.1}
\end{align}
{\color{blue}
and the optimality error
\begin{align}
    \mathbb{E}[R^{k}] &\leq(1-\frac{\alpha\epsilon_{k-1}}{2})\mathbb{E}[R^{k-1}]+NC_1 \epsilon_{k-1}\epsilon_{k-\tau_{k-1}-1}\tau_{k-1}+NC_2 \epsilon_{k-1}\mathbb{E}[S^{k-1}]+NC_2 \epsilon_{k-1}\mathbb{E}[S^{k-\tau_{k-1}-1}]\notag\\
    &\leq(1\hspace{-2pt}-\hspace{-2pt}\frac{\alpha\epsilon_{k-1}}{2})\mathbb{E}[R^{k-1}]\hspace{-2pt}+\hspace{-2pt}NC_1 \epsilon_{k-1}\epsilon_{k-\tau_{k}-1}\tau_{k}\hspace{-2pt}+\hspace{-2pt}NC_2 \epsilon_{k-1}\hspace{-1pt}\mathbb{E}[S^{k-1}]\hspace{-2pt}+\hspace{-2pt}NC_2 \epsilon_{k-1}\mathbb{E}[S^{k-\tau_{k-1}-1}],
    \label{thm:decayepsilon:ineq0.3}
\end{align}
where the second inequality follows from $\tau_{k-1}\leq\tau_k$ and $\epsilon_{k-\tau_{k-1}-1}\leq \epsilon_{k-\tau_{k}-1}$.

Let $i^*$ denote index of the agent with the largest $\rho_i$, i.e. $\rho_{i^*}=\rho$.
For any $k\geq\frac{\rho}{1-\rho}$, Eq.~\eqref{eq:ergodic_consequence}, the definition of the mixing time, and the fact that the max operator is non-expansive imply
\begin{align*}
    \tau_k-\tau_{k-1}&\leq\log_{\rho}(\frac{\epsilon_k}{\hspace{2pt}m_{i^*}})-\log_{\rho}(\frac{\epsilon_{k-1}}{m_{i^*}})=\log_{\rho}(\frac{\epsilon_k}{\epsilon_{k-1}})=\log_{\rho}(\frac{k+1}{k})\leq1.
\end{align*}
Since $\tau_k$, $\tau_{k-1}$ are both integers and $\tau_k\geq\tau_{k-1}$, we have either $\tau_k=\tau_{k-1}$ or $\tau_k=\tau_{k-1}+1$ for all $k\geq\tau_k$.

When $\tau_{k}=\tau_{k-1}$, combining \eqref{thm:decayepsilon:ineq0.1} and \eqref{thm:decayepsilon:ineq0.3} implies $\forall k\geq \tau_k$
\begin{align}
    V^{k}&=\mathbb{E}[R^k] + \mathbb{E}[S^{k}]+\mathbb{E}[S^{k-\tau_k}]\notag\\
    &\leq (1-\frac{\alpha\epsilon_{k-1}}{2})\mathbb{E}[R^{k-1}]+NC_2 \epsilon_{k-1} \mathbb{E}[S^{k-1}]+NC_2 \epsilon_{k-1}\mathbb{E}[S^{k-\tau_{k-1}-1}]+NC_1 \epsilon_{k-1}\epsilon_{k-\tau_k-1}\tau_k\notag\\
    &\hspace{20pt}+\frac{1+\sigma_2^2}{2}\mathbb{E}[S^{k-1}]+\frac{32N B^2\epsilon_{k-1}^2}{1-\sigma_2^2}\mathbb{E}[R^{k-1}]+\frac{NC_0}{1-\sigma_2^2}\epsilon_{k-1}^2+\frac{1+\sigma_2^2}{2}\mathbb{E}[S^{k-\tau_{k-1}-1}]\notag\\
    &\hspace{20pt}+\frac{32N B^2\epsilon_{k-\tau_k-1}^2}{1-\eta^2}\mathbb{E}[R^{k-\tau_{k-1}-1}]+\frac{NC_0}{1-\sigma_2^2}\epsilon_{k-\tau_k-1}^2\notag\\
    &\leq (1-\frac{\alpha\epsilon_{k-1} }{4}+\frac{32N B^2\epsilon_{k-1}^2}{1-\sigma_2^2})\mathbb{E}[R^{k-1}]+\frac{32N B^2\epsilon_{k-\tau_k-1}^2}{1-\eta^2}\mathbb{E}[R^{k-\tau_{k-1}-1}]\notag\\
    &\hspace{20pt}+\left(NC_2\epsilon_{k-1}+\frac{1+\sigma_2^2}{2}\right)\mathbb{E}[S^{k-1}]+\left(NC_2 \epsilon_{k-1}+\frac{1+\sigma_2^2}{2}\right)\mathbb{E}[S^{k-\tau_{k-1}-1}]\notag\\
    &\hspace{20pt}+N(C_1+\frac{2C_0}{1-\sigma_2^2})\epsilon_{k-\tau_k-1}^2\tau_k.
    \label{thm:decayepsilon:ineq7}
\end{align}
Recall from the definition of $S^k$ that $\sqrt{S^k}=\|\bfTheta^k-\1(\btheta^k)^{\top}\|_F$. We can bound the second term in the above inequality using \eqref{lem:theta_k+theta_k-tau:res1} of Lemma \ref{lem:theta_k+theta_k-tau}.
\begin{align*}
    \mathbb{E}[R^{k-\tau_k-1}]&\leq 2\mathbb{E}\left[R^{k-1}+\left\|\btheta^{k-\tau_{k-1}-1}-\btheta^{k-1}\right\|^2\right]\notag\\
    &\leq 2\mathbb{E}\left[R^{k-1}+\left(\frac{1}{2N}\|\btheta^{k-1}\|^2\hspace{-1pt}+\hspace{-1pt}\frac{1}{2N}\sqrt{S^{k-\tau_{k-1}-1}}\hspace{-1pt}+\hspace{-1pt}\frac{1}{2N}\right)^2\right]\notag\\
    &\leq 5\mathbb{E}[R^{k-1}]+3\mathbb{E}[\|\theta^*\|^2]+2\mathbb{E}[S^{k-\tau_{k-1}-1}]+2,
\end{align*}
where the last inequality follows from combining the terms and $N\geq 1$. Using this inequality in \eqref{thm:decayepsilon:ineq7} and simplifying the terms, we get
\begin{align*}
    V^{k}&\leq (1-\frac{\alpha\epsilon_{k-1}}{4}+\frac{192 NB^2\epsilon_{k-\tau_k-1}^2}{1-\sigma_2^2})\mathbb{E}[R^{k-1}]+N(C_1\hspace{-2pt}+\hspace{-2pt}\frac{2C_0}{1-\sigma_2^2}+\frac{32N B^2}{1-\sigma_2^2}(3\mathbb{E}[\|\theta^*\|^2]+2))\epsilon_{k-\tau_k-1}^2\tau_k\notag\\
    &\hspace{10pt}+\left(NC_2\epsilon_{k-1}+\frac{1+\sigma_2^2}{2}\right)\mathbb{E}[S^{k-1}]+\left(\hspace{-1pt}\frac{64 NB^2 \epsilon_{k-\tau_k-1}}{1-\sigma_2^2}\hspace{-2pt}+\hspace{-2pt}N C_2\epsilon_{k-1}\hspace{-1pt}+\hspace{-1pt}\frac{1+\sigma_2^2}{2}\right)\hspace{-2pt}\mathbb{E}[S^{k-\tau_{k-1}-1}].
\end{align*}

Note that by choosing the step size small enough, the terms quadratic in the step size is dominated by terms linear in the step size. Specifically, if the step size sequence satisfies
\begin{align*}
    \epsilon_t\leq\min\{\frac{c_{\tau}\alpha(1-\sigma_2^2)}{768NB^2},\frac{1-\sigma_2^2}{\frac{\alpha}{4}+\frac{128NB^2}{c_{\tau}(1-\sigma_2^2)}+2NC_2}\},\quad\forall t\geq0,
\end{align*}
which is guaranteed by the conditions of Theorem \ref{thm:constant_decay_epsilon_fixedG}, then we can further simplify the above inequality with the help of \eqref{eq:k_tauk} and the fact $\epsilon_{k-\tau_k}\leq \frac{\epsilon_{k}}{c_{\tau}}$ for all $k\geq \tau_k$, which is a result of \eqref{eq:k_tauk}
\begin{align}
    V^{k} &\leq (1-\frac{\alpha\epsilon_{k-1}}{8})\mathbb{E}[R^{k-1}]+N(C_1+\frac{2C_0}{1-\sigma_2^2}+\frac{192 B^2}{1-\sigma_2^2}(3\mathbb{E}[\|\theta^*\|^2]+2))\epsilon_{k-\tau_k-1}^2 \tau_k\notag\\
    &\hspace{10pt}+\left(NC_2\epsilon_{k-1}+\frac{1+\sigma_2^2}{2}\right)\mathbb{E}[S^{k-1}]+\left(\left(\frac{64NB^2}{c_{\tau}(1\hspace{-2pt}-\hspace{-2pt}\sigma_2^2)}+NC_2 \right)\epsilon_{k-1}+\frac{1\hspace{-2pt}+\hspace{-2pt}\sigma_2^2}{2}\right)\mathbb{E}[S^{k-\tau_{k-1}-1}]\notag\\
    &\leq (1-\frac{\alpha\epsilon_{k-1}}{8})\mathbb{E}[R^{k-1}]+\left(1-\frac{\alpha\epsilon_{k-1}}{8}\right)\mathbb{E}[S^{k-1}]+\left(1-\frac{\alpha\epsilon_{k-1}}{8}\right)\mathbb{E}[S^{k-\tau_{k-1}-1}]\notag\\
    &\hspace{10pt}+N(C_1+\frac{2C_0}{1-\sigma_2^2}+\frac{192 B^2}{1-\sigma_2^2}(3\mathbb{E}[\|\theta^*\|^2]+2))\epsilon_{k-\tau_k-1}^2\tau_k\notag\\
    &\leq\left(1-\frac{\alpha\epsilon_{k-1}}{8}\right)V^{k-1}+\frac{N}{1-\sigma_2^2}(C_1+2C_0+192B^2(3\mathbb{E}[\|\theta^*\|^2]+2))\epsilon_{k-\tau_k-1}^2\tau_k\notag\\
    &\leq\left(1-\frac{\alpha\epsilon_{k-1}}{8}\right)V^{k-1}+\frac{13NC_1}{1-\sigma_2^2}\epsilon_{k-\tau_k-1}^2\tau_k,\label{eq:V_case1}
\end{align}
where in the last inequality we combine the constants using the relation
\[C_1+2C_0+192 B^2(3\mathbb{E}[\|\theta^*\|^2]+2)\leq 13 C_1.\]

When $\tau_{k}=\tau_{k-1}+1$, \eqref{thm:decayepsilon:ineq0.1} and \eqref{thm:decayepsilon:ineq0.3} imply $\forall k\geq \tau_k$
\begin{align}
    V^{k}&=\mathbb{E}[R^k] + \mathbb{E}[S^{k}]+\mathbb{E}[S^{k-\tau_k}]\notag\\
    &\leq (1-\frac{\alpha\epsilon_{k-1}}{2})\mathbb{E}[R^{k-1}]+NC_2 \epsilon_{k-1} \mathbb{E}[S^{k-1}]+NC_2 \epsilon_{k-1}\mathbb{E}[S^{k-\tau_{k-1}-1}]+NC_1 \epsilon_{k-1}\epsilon_{k-\tau_k-1}\tau_k\notag\\
    &\hspace{20pt}+\frac{1+\sigma_2^2}{2}\mathbb{E}[S^{k-1}]+\frac{32N B^2\epsilon_{k-1}^2}{1-\sigma_2^2}\mathbb{E}[R^{k-1}]+\frac{NC_0}{1-\sigma_2^2}\epsilon_{k-1}^2+\mathbb{E}[S^{k-\tau_{k-1}-1}]\notag\\
    &\leq (1-\frac{\alpha\epsilon_{k-1} }{4}+\frac{32N B^2\epsilon_{k-1}^2}{1-\sigma_2^2})\mathbb{E}[R^{k-1}]+\left(NC_2 \epsilon_{k-1}+1\right)\left(\mathbb{E}[S^{k-1}]+\mathbb{E}[S^{k-\tau_{k-1}-1}]\right)\notag\\
    &\hspace{20pt}+NC_1 \epsilon_{k-1}\epsilon_{k-\tau_k-1}\tau_k+\frac{NC_0}{1-\sigma_2^2}\epsilon_{k-1}^2\notag\\
    &\leq \left(1+NC_2\epsilon_{k-1}\right)V^{k-1}+\frac{13NC_1}{1-\sigma_2^2}\epsilon_{k-\tau_k-1}^2\tau_k,\label{eq:V_case2}
\end{align}
where the last inequality uses $\epsilon_{k-1}\leq\frac{C_2(1-\sigma_2^2)}{32B^2}$.

Defining
\[\lambda_k= \begin{cases}1-\frac{\alpha\epsilon_{k}}{8}, & \text { if } \tau_{k+1}=\tau_{k} \\ 1+NC_2\epsilon_{k}, & \text { if } \tau_{k+1}=\tau_{k}+1\end{cases}\]
and recursively applying the inequalities \eqref{eq:V_case1} and \eqref{eq:V_case2}, we get for all $k\geq\tau_k$
\begin{align}
    V^k 
    &\leq V^{\tau_k}\prod_{t=\tau_k}^{k-1}\lambda_t+\frac{13NC_1}{1-\sigma_2^2}\sum_{\ell=\tau_k}^{k-1}\epsilon_{\ell-\tau_{\ell}-1}^2 \tau_{\ell} \prod_{t=\ell+1}^{k-1}\lambda_t.
    \label{eq:convergence_generalstepsize}
\end{align}

\eqref{eq:convergence_generalstepsize} characterizes the general convergence behavior under either choice of the step sizes of Theorem \ref{thm:constant_decay_epsilon_fixedG}.
We now specialize to each step size sequence.}

\noindent\textbf{Constant Step Size:} With a constant step size $\epsilon_k=\epsilon$ (hence $\tau_k=\tau_{k-1}=\tau(\epsilon)$ for all $k$), \eqref{eq:convergence_generalstepsize} implies
\begin{align*}
    V^k \hspace{-2pt}&\leq V^{\tau(\epsilon)}\hspace{-4pt}\prod_{\ell=\tau(\epsilon)}^{k-1}\hspace{-2pt}(1\hspace{-2pt}-\hspace{-2pt}\frac{\alpha\epsilon}{8})+\frac{13NC_1}{1-\sigma_2^2}\sum_{\ell=\tau(\epsilon)}^{k-1}\epsilon^2 \tau(\epsilon) \hspace{-2pt}\prod_{t=\ell+1}^{k-1}\hspace{-2pt}(1\hspace{-2pt}-\hspace{-2pt}\frac{\alpha\epsilon}{8})\notag\\
    &\leq V^{\tau(\epsilon)}(1\hspace{-2pt}-\hspace{-2pt}\frac{\alpha\epsilon}{8})^{k-\tau(\epsilon)}\hspace{-2pt}+\hspace{-2pt}\frac{13NC_1}{1-\sigma_2^2} \epsilon^2 \tau(\epsilon) \hspace{-3pt}\sum_{\ell=\tau(\epsilon)}^{k-1}\hspace{-2pt}(1\hspace{-2pt}-\hspace{-2pt}\frac{\alpha\epsilon}{8})^{k-\ell-1}\notag\\
    &\leq V^{\tau(\epsilon)}(1-\frac{\alpha\epsilon}{8})^{k-\tau(\epsilon)}+\frac{13NC_1}{1-\sigma_2^2} \epsilon^2 \tau(\epsilon) \sum_{\ell=0}^{\infty}(1-\frac{\alpha\epsilon}{8})^{\ell}\notag\\
    &\leq V^{\tau(\epsilon)}(1-\frac{\alpha\epsilon}{8})^{k-\tau(\epsilon)}+\frac{104NC_1}{\alpha(1-\sigma_2^2)} \epsilon \tau(\epsilon).
\end{align*} 

To bound $V^{\tau(\epsilon)}$ in terms of the initial conditions, we first bound $S^{\tau(\epsilon)}$. Applying Lemma \ref{lem:consensuserror} recursively, we have
\begin{align}
    S^{\tau(\epsilon)}
    &\leq \frac{1+\sigma_2^2}{2}S^{\tau(\epsilon)-1}+\frac{32N\epsilon^2 B^2}{1-\sigma_2^2}R^{\tau(\epsilon)-1}+\frac{NC_0}{1-\sigma_2^2}\epsilon^2\notag\\
    &\leq S^{\tau(\epsilon)-1}+\frac{32N\epsilon^2 B^2}{1-\sigma_2^2}R^{\tau(\epsilon)-1}+\frac{NC_0}{1-\sigma_2^2}\epsilon^2\notag\\
    &\leq S^{0}+N\epsilon^2\sum_{t=0}^{\tau(\epsilon)-1}\left(\frac{32 B^2}{1-\sigma_2^2}R^{t}+\frac{C_0}{1-\sigma_2^2}\right).
    \label{thm:constantepsilon:ineq7}
\end{align}

By \eqref{lem:theta_k+theta_k-tau:res3} of Lemma \ref{lem:theta_k+theta_k-tau} and the Cauchy-Schwarz inequality,
\begin{align*}
    R^{t}&\leq 2(\|\btheta^{t}-\btheta^0\|^2+R^0)\notag\\
    &\leq 2(\frac{1}{3N}\|\btheta^0\|+\frac{1}{3N}\sqrt{S^0}+\frac{1}{3N})^2+2R^0\notag\\
    &\leq \frac{2}{3N^2}(\|\btheta^0\|^2+S^0+1)+2R^0.
\end{align*}

Then, \eqref{thm:constantepsilon:ineq7} becomes
\begin{align}
    S^{\tau(\epsilon)}
    &\leq S^{0}+N\epsilon^2\sum_{t=0}^{\tau-1}\left(\frac{64 B^2}{3N^2(1-\sigma_2^2)}(\|\btheta^0\|^2+S^0+1)+\frac{64 B^2}{1-\sigma_2^2}R^0+\frac{C_0}{1-\sigma_2^2}\right)\notag\\
    &\leq S^{0}\hspace{-2pt}+\hspace{-2pt}N\epsilon^2\tau(\epsilon)\hspace{-2pt}\left(\frac{128 B^2}{3N^2(1-\sigma_2^2)}(R^0+\|\theta^*\|^2)+\frac{64 B^2}{3N^2(1-\sigma_2^2)}(S^0+1)+\frac{64 B^2}{1-\sigma_2^2}R^0+\frac{C_0}{1-\sigma_2^2}\right)\notag\\
    &\leq(1+\frac{16}{27N^3(1-\sigma_2^2)})S^0+
    \frac{16B^2}{9N(1-\sigma_2^2)}R^0+\frac{C_0}{9NB^2(1-\sigma_2^2)},
    \label{thm:constantepsilon:ineq7.25}
\end{align}
where the last inequality combines the terms with the relations $\epsilon\tau(\epsilon)\leq\frac{1}{6NB}$ and $\tau(\epsilon)\geq 1$.

We then bound 
$R^{\tau(\epsilon)}$.
\begin{align}
    R^{\tau(\epsilon)} &\leq 2\|\btheta^{\tau(\epsilon)}-\btheta^{0}\|^2 + 2R^{0} \notag\\
    &\leq \frac{2}{3N^2}\left(\|\btheta^{0}\|^2+S^{0}+1\right)^2 + 2R^{0}\notag\\
    &\leq (4+\frac{2}{3N^2})\|\btheta^{0}\|^2\hspace{-2pt}+\hspace{-2pt}\frac{2}{3N^2}S^{0}\hspace{-2pt}+\hspace{-2pt}4\|\theta^*\|^2\hspace{-2pt}+\hspace{-2pt}\frac{2}{3N^2},
    \label{thm:constantepsilon:ineq7.5}
\end{align}
where the second inequality again comes from \eqref{lem:theta_k+theta_k-tau:res3} of Lemma \ref{lem:theta_k+theta_k-tau}. Then, we can plug \eqref{thm:constantepsilon:ineq7.25} and \eqref{thm:constantepsilon:ineq7.5} into $V^{\tau(\epsilon)}$
\begin{align*}
    V^{\tau(\epsilon)}&=\mathbb{E}[S^{\tau(\epsilon)}] + \mathbb{E}[S^{\tau(\epsilon)}]+S^{0}\notag\\
    &\leq (4+\frac{2}{3N^2})\|\btheta^{0}\|^2+\frac{2}{3N^2}S^{0}+4\|\theta^*\|^2+\frac{2}{3N^2}+(1+\frac{16}{27N^3(1-\sigma_2^2)})S^0\notag\\
    &\hspace{20pt}+\frac{16}{9N (1-\sigma_2^2)}R^0+\frac{C_0}{9N B^2(1-\sigma_2^2)}+S^{0}\notag\\
    &\leq (2+\frac{16}{27N^3 (1-\sigma_2^2)}+\frac{2}{3N^2})S^0+\frac{2}{3N^2}+\left(4+\frac{2}{3N^2}+\frac{16}{9(1-\sigma_2^2)}\right)\|\btheta^0\|^2\notag\\
    &\hspace{20pt}+\left(4+\frac{16}{9N (1-\sigma_2^2)}\right)\|\theta^*\|^2+\frac{C_0}{9NB^2(1-\sigma_2^2)}\notag\\
    &\leq \frac{4}{1-\sigma_2^2}S^0+\frac{7}{1-\sigma_2^2}\|\btheta^0\|^2+\frac{C_1}{1-\sigma_2^2},
\end{align*}
where the last inequality follows from simplification using $N\geq 1$ and the definition of $C_0$ in \eqref{eq:def_C0}. 

Therefore, for any $i=1,2,...,N$ and any $k>\tau(\epsilon)$,
\begin{align}
    &\mathbb{E}[\|\theta_i^k-\theta^*\|^2] = \mathbb{E}[\|\left(\theta_i^k-\btheta^k\right)+\left(\btheta^k-\theta^*\right)\|^2]\notag\\
    &\leq 2\mathbb{E}[R^k]+2\mathbb{E}[\|\theta_i^k-\btheta^k\|^2]\leq 2\mathbb{E}[R^k]+2\mathbb{E}[S^k]\leq 2V^k\notag\\
    &\leq 2(1-\frac{\alpha\epsilon}{8})^{k-\tau(\epsilon)}V^{\tau(\epsilon)}+\frac{208NC_1}{\alpha(1-\sigma_2^2)}\epsilon\tau(\epsilon)\notag\\
    &\leq (1-\frac{\alpha\epsilon}{8})^{k-\tau(\epsilon)}\Big(\frac{4}{1-\sigma_2^2}S^0+\frac{7\|\btheta^0\|^2}{1-\sigma_2^2}+\frac{C_1}{1-\sigma_2^2}\Big)+\frac{208N\beta C_1 }{\alpha(1-\sigma_2^2)}\epsilon\log(\frac{1}{\epsilon}).
    \label{thm:constantepsilon:ineq8}
\end{align}

\noindent\textbf{Diminishing step size:} 
{\color{blue}
Since $\epsilon_k$ is a non-increasing function in $k$ and $\tau_{t}=\tau_{t-1}+1$ occurs no more than $\tau_k$ times when $t$ increases from $\tau_k$ to $k$, \eqref{eq:convergence_generalstepsize} implies
\begin{align}
    V^k &\leq V^{\tau_k}\prod_{t=\tau_k}^{k-1}\lambda_t+\frac{13NC_1}{1-\sigma_2^2}\sum_{\ell=\tau_k}^{k-1}\epsilon_{\ell-\tau_{\ell}-1}^2 \tau_{\ell} \prod_{t=\ell+1}^{k-1}\lambda_t\notag\\
    &\leq V^{\tau_k}\left(\prod_{t=\tau_k}^{2\tau_k-1}(1+NC_2\epsilon_{t})\right)\left(\prod_{t=2\tau_k}^{k-1}(1-\frac{\alpha\epsilon_{t}}{8})\right)\notag\\
    &\hspace{20pt}+\frac{13NC_1}{1-\sigma_2^2}\sum_{\ell=\tau_k}^{k-1}\epsilon_{\ell-\tau_{\ell}-1}^2 \tau_{\ell} \left(\prod_{t=\ell+1}^{\ell+\tau_k}(1+NC_2\epsilon_{t})\right)\left(\prod_{t=\ell+1+\tau_k}^{k-1}(1-\frac{\alpha\epsilon_{t}}{8})\right).\label{thm:constantepsilon:ineq8.5}
\end{align}

Since $1+x\leq\text{exp}(x)$ for any scalar $x$, we have $\forall\ell\geq\tau_k$,
\begin{align*}
    \left(\prod_{t=\ell}^{\ell-1+\tau_k}(1+NC_2\epsilon_{t})\right)\left(\prod_{t=\ell+\tau_k}^{k-1}(1-\frac{\alpha\epsilon_{t}}{8})\right)&\leq \left(\prod_{t=\ell}^{\ell-1+\tau_k}\exp(NC_2\epsilon_{t})\right)\left(\prod_{t=\ell+\tau_k}^{k-1}\exp(-\frac{\alpha\epsilon_{t}}{8})\right)\notag\\
    &= \left(\exp(\sum_{t=\ell}^{\ell-1+\tau_k}NC_2\epsilon_{t})\right)\left(\exp(-\sum_{t=\ell+\tau_k}^{k-1}\frac{\alpha\epsilon_{t}}{8})\right)\notag\\
    &\leq \left(\exp(NC_2\epsilon_{\tau_k}\tau_k)\right)\left(\exp(-\sum_{t=\ell+\tau_k}^{k-1}\frac{\alpha\epsilon}{8(t+1)})\right)\notag\\
    &\leq2\exp(-\sum_{t=\ell+\tau_k}^{k-1}\frac{\alpha\epsilon}{8(t+1)}),
\end{align*}
where the third inequality use $NC_2\epsilon_{\tau_k}\tau_k\leq1/2\leq\log(2)$ afforded by the step size condition.
Note that \[\sum_{t=k_1}^{k_2}\frac{1}{t+1}\geq \int_{k_1}^{k_2+1}\frac{1}{t+1} dt=\log(\frac{k_2+2}{k_1+1}),\]
which simplifies the inequality above
\begin{align*}
    \left(\prod_{t=\ell}^{\ell-1+\tau_k}(1+NC_2\epsilon_{t})\right)\left(\prod_{t=\ell+\tau_k}^{k-1}(1-\frac{\alpha\epsilon_{t}}{8})\right)\leq2\exp(-\sum_{t=\ell+\tau_k}^{k-1}\frac{\alpha\epsilon}{8(t+1)})\leq \frac{2(\ell+\tau_k+2)}{k+1}
\end{align*}
if $\alpha\epsilon\geq8$.
Using this bound in \eqref{thm:constantepsilon:ineq8.5}, we have
\begin{align}
    V^k &\leq \frac{4(\tau_k\hspace{-2pt}+\hspace{-2pt}1)}{k+1}V^{\tau_k}\hspace{-2pt}+\hspace{-2pt}\frac{13NC_1}{1-\sigma_2^2}\sum_{\ell=\tau_k}^{k-1}\epsilon_{\ell-\tau_{\ell}-1}^2 \tau_{\ell} \frac{2(\ell+\tau_k+1)}{k+1}\notag\\
    &\leq\frac{4(\tau_k+1)}{k+1}V^{\tau_k}+\frac{52NC_1}{1-\sigma_2^2}\tau_{k-1}\sum_{\ell=\tau_k}^{k-1}\epsilon_{\ell-\tau_{\ell}-1}^2 \frac{\ell+1}{k+1}\notag\\
    &\leq\frac{4(\tau_k+1)}{k+1}V^{\tau_k}+\frac{52NC_1\tau_{k-1}}{c_{\tau}^2(1-\sigma_2^2)(k+1)}\sum_{\ell=\tau_k}^{k-1}\frac{\ell+1}{\ell^2}\notag\\
    &\leq\frac{4(\tau_k+1)}{k+1}V^{\tau_k}+\frac{208NC_1\tau_{k-1}}{c_{\tau}^2(1-\sigma_2^2)(k+1)}\sum_{\ell=\tau_k}^{k-1}\frac{1}{\ell+1},
    \label{thm:decayeps:eq1}
\end{align}
where the third inequality uses $\epsilon_{k-\tau_k}\leq \frac{\epsilon_{k}}{c_{\tau}}$, for all $k\geq \tau_k$, which is a result of \eqref{eq:k_tauk}, and the fourth inequality follows from $\frac{\ell+1}{\ell^2}\leq\frac{4}{\ell+1}$ for all $\ell\geq1$. Applying
\[\sum_{t=k_1}^{k_2}\frac{1}{t+1}\leq \int_{k_1-1}^{k_2}\frac{1}{t+1} dt=\log(\frac{k_2+1}{k_1})\]
in \eqref{thm:decayeps:eq1}, we have
\begin{align*}
    V^k 
    &\leq \frac{4(\tau_k+1)}{k+1}V^{\tau_k}+\frac{208NC_1\tau_{k-1}}{c_{\tau}^2(1-\sigma_2^2)(k+1)}\log(\frac{k}{\tau_k})\notag\\
    &\leq\frac{4(\beta(\log(k+1)-\log(\epsilon))+1)}{k+1}V^{\tau_k}+\frac{208NC_1\beta(\log(k)-\log(\epsilon))\log(k)}{c_{\tau}^2(1-\sigma_2^2)(k+1)}.
\end{align*}
}

We can bound $V^{\tau_k}$ by the initial conditions the same way as in the case of a constant step size and get 
\begin{align*}
    V^{\tau_k}
    &\leq \frac{4}{1-\sigma_2^2}S^0+\frac{7}{1-\sigma_2^2}\|\btheta^0\|^2+\frac{C_1}{1-\sigma_2^2}.
\end{align*}
Then, similar to \eqref{thm:constantepsilon:ineq8}, for any $i\in=1,2,...,N$ and any $k>\tau_k$,
\begin{align*}
    &\mathbb{E}[\|\theta_i^k-\theta^*\|^2]\leq 2V^k\notag\\
    &\leq \frac{4(\beta(\log(k+1)-\log(\epsilon))+1)}{(1-\sigma_2^2)(k+1)}\Big(4S^0+7\|\btheta^0\|^2+C_1\Big)+\frac{208N\beta C_1\log(k)(\log(k)-\log(\epsilon))}{c_{\tau}^2 (1-\sigma_2^2)(k+1)}.
\end{align*}
\qed

{\color{blue}
\section{CONCLUSION}
In this work, we study the finite-time convergence of decentralized stochastic approximation over a network of agents, where the data at each agent are generated from a Markov process. We remove the restrictive assumptions on bounded updates that are usually made in the existing literature. We discuss and numerically solve two problems in distributed optimization and multi-task learning that can be viewed as special cases of our framework. Future directions from this work include studying the {\sf DCSA} framework under time-varying and/or directed communication graphs, communication constraints, and asynchronous communication.
}

\appendix


\bibliographystyle{IEEEtran}
\bibliography{references}

\section{Proof of Technical Lemmas}

\subsection{Proof of Lemma \ref{lem:B_bounded}}
\label{sec:B_bounded}

Using Assumption \ref{assump:F_Lipschitz},
\begin{align*}
    \|F_i(X_i,\theta)\|\hspace{-3pt}-\hspace{-3pt}\|F_i(X_i,0)\|\leq \|F_i(X_i,\theta)\hspace{-3pt}-\hspace{-3pt}F_i(X_i,0)\|\leq L\|\theta\|.
\end{align*}

Re-arranging terms, we have for all $i$ and $X_i$
\begin{align*}
    \|F_i(X_i,\theta)\|&\leq L\|\theta\|+\|F_i(X_i,0)\|\leq B\|\theta\|+B\leq B(\|\theta\|+1).
\end{align*}

The same argument extends easily to $\|\bar{F}_i(\theta)\|$.

\subsection{Proof of Lemma \ref{lem:theta_k+theta_k-tau}}
\label{sec:proofs:theta_k+theta_k-tau}




As a result of Lemma \ref{lem:B_bounded},
\begin{align}
    \|\btheta^{k+1}\|-\|\btheta^{k}\|&\leq \|\btheta^{k+1}-\btheta^{k}\|\notag\\
    &=\|\frac{\epsilon_k}{N}\sum_{i=1}^{N}F_i(X_{i}^{k}, \theta_i^{k})\|\notag\\
    &\leq \frac{\epsilon_k}{N}\sum_{i=1}^{N}B(\|\theta_i^k\|+1)\notag\\
    &\leq \frac{\epsilon_k B}{N}\sum_{i=1}^{N}(\|\btheta^k\|+\|\theta_i^k-\btheta^k\|+1)\notag\\
    &\leq \epsilon_k B (\|\btheta^k\|+\sqrt{S^k}+1).
    \label{lem:theta_k+theta_k-tau:ineq0}
\end{align}
The last inequality of \eqref{lem:theta_k+theta_k-tau:ineq0} follows from the fact that \[\|\theta_i^k-\btheta^k\|\leq\sqrt{S^k},\] 
which is a simple consequence of the definition of $S^k$ in \eqref{eq:def_S}. Re-arranging the terms in \eqref{lem:theta_k+theta_k-tau:ineq0} leads to
\begin{align*}
    \|\btheta^{k+1}\|&\leq(1+\epsilon_k B)\|\btheta^{k}\|+\epsilon_k B (\sqrt{S^k}+1).
\end{align*}

The matrix form of the update rule \eqref{Alg:DCSA:Update} is
\begin{align}
\bfTheta^{k+1} = \Wbf\bfTheta^{k} + \epsilon_k \mF(\mX^{k},\bfTheta^{k}).
\label{eq:update_thetavector_decayepsilon}
\end{align}

\eqref{eq:update_thetavector_decayepsilon} implies that the update for $\btheta^k$ observes
\begin{align}
    \btheta^{k+1} = \frac{1}{N}(\bfTheta^{k})^{\top}\1 + \frac{\epsilon_k}{N} \mF(\mX^k,\bfTheta^k)^{\top}\1.
\label{eq:update_thetabar_decayepsilon}
\end{align}

Using the update rule of $\bfTheta^{k+1}$ and $\btheta^{k+1}$,
\begin{align}
\bfTheta^{k+1} - \1(\bar{\theta}^{k+1})^{\top}
&=\Wbf\bfTheta^{k} - \frac{1}{N}\1\1^{\top} \bfTheta^{k}+\epsilon_k\left(\mF(\mX^{k},\bfTheta^{k}) - \frac{1}{N}\1\1^{\top}\mF(\mX^{k},\bfTheta^{k})\right)\notag\\
&=(\Wbf- \frac{1}{N}\1\1^{\top})\bfTheta^{k} \hspace{-2pt}-\hspace{-2pt} (\Wbf \1\hspace{-2pt}-\hspace{-2pt}\frac{1}{N}\1\1^{\top}\1 )(\btheta^{k})^{\top}\hspace{-2pt}+\hspace{-2pt}\epsilon_k(\mF(\mX^{k},\bfTheta^{k}) \hspace{-2pt}-\hspace{-2pt} \frac{1}{N}\1\1^{\top}\mF(\mX^{k},\bfTheta^{k}))\notag\\
&=\Big(\Wbf- \frac{1}{N}\1\1^{\top}\Big)\Big(\bfTheta^{k} - \1(\btheta^{k})^{\top}\Big)+\epsilon_k\left(\Ibf - \frac{1}{N}\1\1^{\top}\right)\mF(\mX^{k},\bfTheta^{k}).
\label{eq:consensuserror_equation}
\end{align}

Note that for any matrix $\Ybf\in\mathbb{R}^{N\times N},\Zbf\in\mathbb{R}^{N\times d}$, we have $\|\Ybf\Zbf\|_F\leq\|\Ybf\|_* \|\Zbf\|_F$, since
\begin{align*}
\|\Ybf\Zbf\|_F^2 = \sum_{i=1}^{N}\|\Ybf z_i\|^2 \leq \sum_{i=1}^{N}\|\Ybf\|_{*}^2\|z_{i}\|^2\leq \|\Ybf\|_{*}^2\|\sum_{i=1}^{N}\|z_{i}\|^2=\|\Ybf\|_{*}^2\|\Zbf\|_F^2,
\end{align*}
where $z_i$ denotes the $i_{\text{th}}$ column of $\Zbf$. This inequality implies
\begin{align*}
\left\|\bfTheta^{k+1} - \1(\bar{\theta}^{k+1})^{\top}\right\|_F &\leq \left\|\Wbf-\frac{1}{N}\1\1^{\top}\right\|_*\hspace{-5pt}\sqrt{S^{k}} \hspace{-1pt}+\hspace{-1pt}\epsilon_k \left\|\Ibf-\frac{1}{N}\1\1^{\top}\right\|_* \left\| \mF(\mX^{k},\bfTheta^{k})\right\|_F \notag\\
&\leq \sqrt{S^{k}} + \epsilon_k\sum_{i=1}^{N}\|F_i(X^k,\theta_i^k)\|\notag\\
&\leq \sqrt{S^{k}} + \epsilon_k\sum_{i=1}^{N}B(\|\theta_i^k\|+1)\notag\\
&\leq \sqrt{S^{k}} + \epsilon_k B\sum_{i=1}^{N}(\|\theta_i^k-\bar{\theta}^k\|+\|\bar{\theta}^k\|+1)\notag\\
&\leq \left(1+\epsilon_k NB\right)\sqrt{S^{k}} + \epsilon_k NB\|\bar{\theta}^{k}\| + \epsilon_k NB,
\end{align*}
where the first inequality follows from $\|\Wbf-\frac{1}{N}\1\1^{\top}\|_*\leq1$ and $\|\Ibf-\frac{1}{N}\1\1^{\top}\|_*\leq1$.

Define $y^k = \|\btheta^{k}\|+\sqrt{S^{k}}$. Using the two inequalities above,
\begin{align*}
    &y^{k+1} = \|\btheta^{k+1}\|+\sqrt{S^{k+1}}\notag\\
    &\leq (1+\epsilon_k B)\|\btheta^{k}\|+\epsilon_k B (\sqrt{S^k}+1)+(1+\epsilon_k NB)\sqrt{S^k}+\epsilon_k NB\|\btheta^k\|+\epsilon_k NB\notag\\
    &\leq(1+2\epsilon_k NB)\|\btheta^{k}\|+(1+2\epsilon_k NB)\sqrt{S^k}+2\epsilon_k NB\notag\\
    &=(1+2N\epsilon_k B)y^k+2\epsilon_k NB.
\end{align*}

First, we show the bound for $k\geq\tau_k$. For $t'\in[k-\tau_k,k]$,
\begin{align}
    y^{t'}&\leq(1+2\epsilon_{k-\tau_k} N B)^{t'-k+\tau_k}y^{k-\tau_k}+2\epsilon_{k-\tau_k} NB\sum_{u=k-\tau_k}^{t'-1} (1+2\epsilon_{k-\tau_k} NB)^{t'-u-1}\notag\\
    &\leq (1+2\epsilon_{k-\tau_k} N B)^{\tau_k} y^{k-\tau_k}+2\epsilon_{k-\tau_k} NB\tau_k(1+2\epsilon_{k-\tau_k} N B)^{\tau_k}\notag\\
    &\leq e^{2\epsilon_{k-\tau_k} NB\tau_k}y^{k-\tau_k}+2\epsilon_{k-\tau_k} NB\tau_ke^{2\epsilon_{k-\tau_k} NB\tau_k}\notag\\
    &\leq 2y^{k-\tau_k}+4\epsilon_{k-\tau_k} NB\tau_k,
    \label{lem:theta_k+theta_k-tau:ineq3}
\end{align}
where we have used the fact that $1+x\leq e^{x}$ for $x\geq 0$ and the relationship $e^{2\epsilon_{k-\tau_k} NB\tau_k}\leq 2$, which comes from the condition on the step size $2\epsilon_{k-\tau_k}\tau_kNB\leq\frac{1}{3}\leq\text{log}(2)$. Then,
\begin{align}
    \|\btheta^k-\btheta^{k-\tau_k}\|\hspace{-2pt}&\leq \sum_{t'=k-\tau_k}^{k-1}\|\btheta^{t'+1}-\btheta^t\|\notag\\
    &\leq \epsilon_{k-\tau_k} B \hspace{-5pt}\sum_{t'=k-\tau_k}^{k-1}\hspace{-5pt} (\|\btheta^{t'}\|+\sqrt{S^{t'}}+1)\notag\\
    &\leq \epsilon_{k-\tau_k} B \hspace{-5pt}\sum_{t'=k-\tau_k}^{k-1}\hspace{-5pt} (y^{t'}+1)\notag\\
    &\leq \epsilon_{k-\tau_k} B\hspace{-5pt} \sum_{t'=k-\tau_k}^{k-1}\hspace{-5pt} (2y^{k-\tau_k}+4\epsilon_{k-\tau_k} NB\tau_k+1)\notag\\
    &\leq \epsilon_{k-\tau_k} B \tau_k \left(2\|\btheta^{k-\tau_k}\|+2\sqrt{S^{k-\tau_k}}+2\right)\notag\\
    &\leq 2\epsilon_{k-\tau_k} B \tau_k\|\btheta^k-\btheta^{k-\tau_k}\|+2\epsilon_{k-\tau_k} B \tau_k\|\btheta^k\|+2\epsilon_{k-\tau_k} B \tau_k\sqrt{S^{k-\tau_k}}\hspace{-2pt}+\hspace{-2pt}2\epsilon_{k-\tau_k} B\tau_k,
    \label{lem:theta_k+theta_k-tau:ineq4}
\end{align}
where the second inequality is from \eqref{lem:theta_k+theta_k-tau:ineq0}, and the second last inequality comes from the step size rule $\epsilon_{k-\tau_k}\tau_k\leq \frac{1}{6NB}$.

Re-arranging terms and again using the condition $\epsilon_{k-\tau_k}\tau_k\leq\frac{1}{6B}$, we have
\begin{align*}
    &\|\btheta^k-\btheta^{k-\tau_k}\|\notag\\
    &\leq 3\epsilon_{k-\tau_k} B\tau_k\|\btheta^k\|+3\epsilon_{k-\tau_k} B \tau_k\sqrt{S^{k-\tau_k}}+3\epsilon_{k-\tau_k} B\tau_k\notag\\
    &\leq \frac{1}{2N}\|\btheta^k\|+\frac{1}{2N}\sqrt{S^{k-\tau_k}}+\frac{1}{2N}.
\end{align*}


Now it remains to show the bound on $\|\btheta^k-\btheta^0\|$ for $k\leq\tau_k$. Similar to \eqref{lem:theta_k+theta_k-tau:ineq3}, for $t'\in\left[0,k\right]$ we have
\begin{align*}
    y^{t'}\leq 2y^0+4\epsilon_0 NB\tau_k=2y^0+4\epsilon NB\tau_k,
\end{align*}
which implies for all $k\leq\tau_k$
\begin{align*}
    \|\btheta^k-\btheta^{0}\|&\leq \sum_{t'=0}^{k-1}\|\btheta^{t'+1}-\btheta^{t'}\|\notag\\
    &\leq \epsilon B \sum_{t'=0}^{k-1} (\|\btheta^{t'}\|+\sqrt{S^{t'}}+1)\notag\\
    &= \epsilon B \sum_{t'=0}^{k-1} (y^{t'}+1)\notag\\
    &\leq \epsilon B \sum_{t'=0}^{k-1} (2y^{0}+4\epsilon NB\tau_k+1)\notag\\
    &\leq \epsilon B \tau_k \left(2\|\btheta^{0}\|+2\sqrt{S^{0}}+2\right)\leq \frac{1}{3N}(\|\btheta^{0}\|+\sqrt{S^{0}}+1),
\end{align*}
where again the second inequality comes from \eqref{lem:theta_k+theta_k-tau:ineq0} and the last inequality follows from the condition on the step size.

\qed

\subsection{Proof of Lemma \ref{lem:consensuserror}}


We have shown in \eqref{eq:consensuserror_equation} that for all $k\geq 0$
\begin{align}
\bfTheta^k - \1(\btheta^k)^{\top}&=(\Wbf-\frac{1}{N}\1\1^{\top})\left(\bfTheta^{k-1}-\1(\btheta^{k-1})^{\top}\right)+\epsilon_{k-1}\left(\Ibf-\frac{1}{N}\1\1^{\top}\right)\mF(X^{k-1},\bfTheta^{k-1}).
\label{thm:decayepsilon:ineq2.25}
\end{align}
As $\|\Wbf-\frac{1}{N}\1\1^{\top}\|_*\leq\sigma_2$ and $\|\Ibf-\frac{1}{N}\1\1^{\top}\|_*\leq1$, we have
\begin{align}
S^k&=\|\bfTheta^k - \1(\btheta^{k})^{\top}\|_F^2\notag\\
&= \left\|(\Wbf-\frac{1}{N}\1\1^{\top}) \left(\bfTheta^{k-1}-\1(\btheta^{k-1})^{\top}\right)+\epsilon_{k-1}\left(\Ibf-\frac{1}{N}\1\1^{\top}\right)\mF(X^{k-1},\bfTheta^{k-1})\right\|_F^2\notag\\
&\leq \left(1+\frac{1-\sigma_2^2}{4\sigma_2^2}\right)\Big\|(\Wbf-\frac{1}{N}\1\1^{\top}) \left(\bfTheta^{k-1}-\1(\btheta^{k-1})^{\top}\right)\Big\|_F^2 \notag\\
&\hspace{20pt}+ \left(1+\frac{8\sigma_2^2}{1-\sigma_2^2}\right)\hspace{-3pt} \Big\|\epsilon_{k-1}\left(\Ibf-\frac{1}{N}\1\1^{\top}\right)\hspace{-3pt}\mF(X^{k-1},\bfTheta^{k-1})\hspace{-1pt}\Big\|_F^2\notag\\
&\leq \sigma_2^2 \left(1+\frac{1-\sigma_2^2}{4\sigma_2^2}\right)S^{k-1} + (1+\frac{8\sigma_2^2}{1-\sigma_2^2})\epsilon_{k-1}^2 \| \mF(\mX^{k-1},\bfTheta^{k-1})\|_F^2\notag\\
&= \frac{1+3\sigma_2^2}{4}S^{k-1}+\frac{(1+7\sigma_2^2)\epsilon_{k-1}^2}{1-\sigma_2^2}\sum_{i=1}^{N}\| F_i(X_i^{k-1},\theta_i^{k-1})\|^2\notag\\
&\leq \frac{1+3\sigma_2^2}{4}S^{k-1}+\frac{8\epsilon_{k-1}^2}{1-\sigma_2^2}\sum_{i=1}^{N}B^2(\|\theta_i^{k-1}\|+1)^2,
\label{thm:decayepsilon:ineq2.5}
\end{align}
where in the first inequality we use the fact that $\|a+b\|^2\leq(1+\frac{1}{\eta})\|a\|^2+(1+2\eta)\|b\|^2$ for all vectors $a,b$ and $\eta>0$. The third inequality follows from Lemma \ref{lem:B_bounded} and $\sigma_2^2\leq 1$. By Cauchy-Schwarz inequality we have 
\begin{align*}
    (\|\theta_i^{k-1}\|+1)^2&\leq (\|\theta_i^{k-1}-\bar{\theta}^{k-1}\|+\|\bar{\theta}^{k-1}-\theta^*\|+\|\theta^*\|+1)^2\notag\\
    &\leq 4(\|\theta_i^{k-1}-\bar{\theta}^{k-1}\|^2+R^{k-1}+\|\theta^*\|^2+1),
\end{align*}

Applying this to \eqref{thm:decayepsilon:ineq2.5} implies
\begin{align*}
S^k&\leq \frac{1+3\sigma_2^2}{4}S^{k-1}+\frac{8\epsilon_{k-1}^2}{1-\sigma_2^2}\sum_{i=1}^{N}B^2(\|\theta_i^{k-1}\|+1)^2\notag\\
&\leq \frac{1+3\sigma_2^2}{4}S^{k-1}+\frac{32\epsilon_{k-1}^2 B^2}{1-\sigma_2^2}\sum_{i=1}^{N}(\|\theta_i^{k-1}-\bar{\theta}^{k-1}\|^2+R^{k-1}+\|\theta^*\|^2+1)\notag\\
&\leq (\frac{1+3\sigma_2^2}{4}+\frac{32\epsilon_{k-1}^2 B^2 N}{1-\sigma_2^2})S^{k-1}+\frac{32\epsilon_{k-1}^2 B^2 N}{1-\sigma_2^2}R^{k-1}+\frac{NC_0}{1-\sigma_2^2}\epsilon_{k-1}^2\notag\\
&\leq \frac{1+\sigma_2^2}{2}S^{k-1}+\frac{32\epsilon_{k-1}^2 B^2 N}{1-\sigma_2^2}R^{k-1}+\frac{NC_0}{1-\sigma_2^2}\epsilon_{k-1}^2,
\end{align*}
In the second last inequality, we define $C_0=16B^2\left(\|\theta^*\|^2+1\right)$. The last inequality is a result of the step size condition $\epsilon_{k-1}\leq \frac{1-\sigma_2^2}{8\sqrt{2}BN}$.

\qed

\subsection{Proof of Lemma \ref{lem:optimalityerror}}

To show Lemma \ref{lem:optimalityerror}, we introduce the following technical lemmas which bounds the cross term under Markovian data. 

\begin{lem}
We have for all $k\geq\tau_k$
\begin{align*}
    \mathbb{E}[\langle\btheta^k-\theta^*,\sum_{i=1}^{N}\left(F_i(X_i^k,\theta_i^k)-\bar{F}_i(\btheta^k)\right)\rangle]&\leq \frac{\alpha}{2}\mathbb{E}[R^k]+M_1\epsilon_{k-\tau_k}\tau_k\mathbb{E}[R^k]+M_2\mathbb{E}[S^k]\\
    &\hspace{20pt}+M_3\mathbb{E}[S^{k-\tau_k}]+\epsilon_{k-\tau_k}\tau_k M_4\left(\|\theta^*\|^2+1\right),
\end{align*}
where 
\begin{align*}
M_1&=(\frac{45B}{4}+30B^2+48BL)N,\\
M_2&=\frac{5B}{N}+\frac{NL^2}{\alpha}+\frac{L}{2},\\ 
M_3&=5B+\frac{5}{12N}+\frac{4NL^2}{\alpha}+5L,\\ 
M_4&=(\frac{45B}{4}+30B^2+45BL)N.
\end{align*}
\label{lem:innerproduct}
\end{lem}

\textbf{Proof (of Lemma \ref{lem:optimalityerror}):}
From the update rule \eqref{Alg:DCSA:Update},
\begin{align}
R^{k+1}&\leq \|\bar{\theta}^{k}+\frac{\epsilon_k}{N}\sum_{i=1}^{N}F_i(X_{i}^{k}, \theta_i^{k})-\theta^*\|^2\notag\\
&= R^{k}+\|\frac{\epsilon_k}{N}\sum_{i=1}^{N}F_i(X_{i}^{k}, \theta_i^{k})\|^2+2\epsilon_k\langle\btheta^k-\theta^*,\sum_{i=1}^{N}\bar{F}_i(\btheta^k)\rangle\notag\\
&\hspace{20pt}+2\epsilon_k\langle\btheta^k-\theta^*,\sum_{i=1}^{N}\left(F_i(X_i^k,\theta_i^k)-\bar{F}_i(\btheta^k)\right)\rangle\notag\\
&= R^{k}+\|\frac{\epsilon_k}{N}\sum_{i=1}^{N}F_i(X_{i}^{k}, \theta_i^{k})\|^2+2\epsilon_k\langle\btheta^k-\theta^*,\sum_{i=1}^{N}\bar{F}_i(\btheta^k)-\sum_{i=1}^{N}\bar{F}_i(\theta^*)\rangle\notag\\
&\hspace{20pt}+2\epsilon_k\langle\btheta^k-\theta^*,\sum_{i=1}^{N}\left(F_i(X_i^k,\theta_i^k)-\bar{F}_i(\btheta^k)\right)\rangle\notag\\
&\leq R^{k}+\|\frac{\epsilon_k}{N}\sum_{i=1}^{N}F_i(X_{i}^{k}, \theta_i^{k})\|^2-2\alpha\epsilon_kR^k+2\epsilon_k\langle\btheta^k-\theta^*,\sum_{i=1}^{N}\left(F_i(X_i^k,\theta_i^k)-\bar{F}_i(\btheta^k)\right)\rangle\notag\\
&= (1-2\alpha\epsilon_k)R^{k}+\|\frac{\epsilon_k}{N}\sum_{i=1}^{N}F_i(X_{i}^{k}, \theta_i^{k})\|^2+2\epsilon_k\langle\btheta^k-\theta^*,\sum_{i=1}^{N}\left(F_i(X_i^k,\theta_i^k)-\bar{F}_i(\btheta^k)\right)\rangle,
\label{thm:decayepsilon:ineq1}
\end{align}
where the first equality follows from $\sum_{i=1}^{N}\bar{F}_i(\theta^*)=0$, and the third inequality is a result of Assumption \ref{assump:Fbar_stronglymonotone}.

We bound the second term of \eqref{thm:decayepsilon:ineq1} with Lemma \ref{lem:B_bounded}
\begin{align}
    \|\frac{\epsilon_k}{N}\sum_{i=1}^{N}F_i(X_{i}^{k}, \theta_i^{k})\|^2&\leq\frac{\epsilon_k^2}{N}\sum_{i=1}^{N}\|F_i(X_i^k,\theta_i^k)\|^2\notag\\
    &\leq\frac{\epsilon_k^2}{N}\sum_{i=1}^{N}B^2(\|\theta_i^k\|+1)^2\notag\\
    &\leq \frac{\epsilon_k^2 B^2}{N}\sum_{i=1}^{N}(3\|\btheta^k\|^2+3\|\theta_i^k-\btheta^k\|^2+3)\notag\\
    &\leq 3\epsilon_k^2 B^2(\|\btheta^k\|^2+1)+\frac{3\epsilon_k^2 B^2}{N}S^k\notag\\
    &\leq 6\epsilon_k^2 B^2(R^k+\|\theta^*\|^2+1)+\frac{3\epsilon_k^2 B^2}{N}S^k.
    \label{thm:decayepsilon:ineq1.5}
\end{align}

Lemma \ref{lem:innerproduct} bounds the third term of \eqref{thm:decayepsilon:ineq1}. This, along with \eqref{thm:decayepsilon:ineq1.5} and \eqref{thm:decayepsilon:ineq1}, implies that
\begin{align*}
    \mathbb{E}[R^{k+1}] &\leq \mathbb{E}\left[(1-2\alpha\epsilon_k)R^{k}+\|\frac{\epsilon_k}{N}\sum_{i=1}^{N}F_i(X_{i}^{k}, \theta_i^{k})\|^2+2\epsilon_k\langle\btheta^k-\theta^*,\sum_{i=1}^{N}\left(F_i(X_i^k,\theta_i^k)-\bar{F}_i(\btheta^k)\right)\rangle\right] \notag\\
    &\leq (1-2\alpha\epsilon_k)\mathbb{E}[R^k]+6\epsilon_k^2 B^2(\mathbb{E}[R^k]+\|\theta^*\|^2+1)+\frac{3\epsilon_k^2 B^2}{N}\mathbb{E}[S^k]+\alpha\epsilon_k\mathbb{E}[R^k]\notag\\
    &\hspace{20pt}+2M_1\epsilon_k\epsilon_{k-\tau_k}\tau_k\mathbb{E}[R^k]+2M_2\epsilon_k\mathbb{E}[S^k]+2M_3\epsilon_k\mathbb{E}[S^{k-\tau_k}]+2\epsilon_k\epsilon_{k-\tau_k}\tau_k M_4\left(\|\theta^*\|^2+1\right)\notag\\
    &\leq(1-\alpha\epsilon_k)\mathbb{E}[R^k]+\left(2M_1+6B^2\right)\epsilon_k\epsilon_{k-\tau_k}\tau_k\mathbb{E}[R^k]+\left(\frac{B}{2N^2}+2M_2\right) \epsilon_k\mathbb{E}[S^k]\notag\\
    &\hspace{20pt}+2M_3 \epsilon_k\mathbb{E}[S^{k-\tau_k}]+NC_1 \epsilon_k\epsilon_{k-\tau_k}\tau_k\notag\\
    &\leq(1-\frac{\alpha\epsilon_k}{2})\mathbb{E}[R^k]+NC_1 \epsilon_k\epsilon_{k-\tau_k}\tau_k+N\left(\frac{21B}{2}+\frac{2L^2}{\alpha}+L\right) \epsilon_k\mathbb{E}[S^k]\notag\\
    &\hspace{20pt}+N\left(10B+\frac{5}{6}+\frac{8L^2}{\alpha}+10L\right) \epsilon_k\mathbb{E}[S^{k-\tau_k}]\notag\\
    &\leq(1-\frac{\alpha\epsilon_k}{2})\mathbb{E}[R^k]+NC_1 \epsilon_k\epsilon_{k-\tau_k}\tau_k+NC_2 \epsilon_k\mathbb{E}[S^k]+NC_2 \epsilon_k\mathbb{E}[S^{k-\tau_k}],
\end{align*}
where the third inequality simply follows from re-arranging/grouping of the terms. In the fourth inequality, we use the condition that $\epsilon_{k-\tau_k}\tau_k\leq\frac{\alpha}{45NB+132NB^2+192NBL}$, and $C_{1}$ and $C_{2}$ are defined in \eqref{notation:constant_C}.

\qed

\subsection{Proof of Lemma \ref{lem:innerproduct}}
Define
\begin{align}
    &\Tcal_1^k=\langle\btheta^k-\btheta^{k-\tau_k},\sum_{i=1}^{N}\left(F_i(X_i^k,\theta_i^k)-\bar{F}_i(\btheta^k)\right)\rangle,\notag\\
    &\Tcal_2^k=\langle\btheta^{k-\tau_k}-\theta^*,\sum_{i=1}^{N}\left(F_i(X_i^k,\theta_i^{k-\tau_k})-\bar{F}_i(\theta_i^{k-\tau_k})\right)\rangle,\notag\\
    &\Tcal_3^k=\langle\btheta^{k-\tau_k}-\theta^*,\sum_{i=1}^{N}\left(F_i(X_i^k,\theta_i^k)-F_i(X_i^k,\theta_i^{k-\tau_k})\right)\rangle,\notag\\
    &\Tcal_4^k=\langle\btheta^{k-\tau_k}-\theta^*,\sum_{i=1}^{N}\left(\bar{F}_i(\theta_i^{k-\tau_k})-\bar{F}_i(\btheta^{k-\tau_k})\right)\rangle,\notag\\
    &\Tcal_5^k=\langle\btheta^{k-\tau_k}-\theta^*,\sum_{i=1}^{N}\left(\bar{F}_i(\btheta^{k-\tau_k})-\bar{F}_i(\btheta^k)\right)\rangle.
    \label{eq:T_def}
\end{align}

We introduce the following two lemmas needed in the proof of Lemma \ref{lem:innerproduct}, and note that they are essentially consequences of the Lipschitz continuity of $F_i$ and $\bar{F}_i$.

\begin{lem}
We have for all $k\geq\tau_k$
\begin{align*}
    \Tcal_1^k\hspace{-2pt}&\leq\hspace{-2pt} 30\epsilon_{k-\tau_k}\tau_k B^2\hspace{-2pt}\left(\hspace{-2pt}NR^k+NS^{k-\tau_k}+S^k+N (\|\theta^*\|^2+1)\hspace{-1pt}\right).
\end{align*}
\label{lem:innerproduct_term1}
\end{lem}
\begin{lem}
We have for all $k\geq\tau_k$
\begin{align*}
    \Tcal_3^k\hspace{-2pt}+\hspace{-2pt}\Tcal_4^k\hspace{-2pt}+\hspace{-2pt}\Tcal_5^k&\leq 48\epsilon_{k-\tau_k}\tau_k N BLR^k+\frac{\alpha}{2}R^k+\frac{45\epsilon_{k-\tau_k}\tau_k NBL}{2}(2\|\theta^*\|^2+1)\notag\\
    &\hspace{20pt}\hspace{-2pt}+\hspace{-2pt}(\frac{NL^2}{\alpha}+\frac{L}{2})S^{k}+(\frac{4NL^2}{\alpha}+5L)S^{k-\tau_k}.
\end{align*}
\label{lem:innerproduct_term345}
\end{lem}

\textbf{Proof (of Lemma \ref{lem:innerproduct}):}

To simplify notations, we define the norm of the local consensus error $r_i^k \hspace{-2pt}=\hspace{-2pt} \|\theta_i^k-\btheta^k\|$. Note that $r_i^k$ relates to $S^k$
\begin{align*}
    \sum_{i=1}^{N} (r_i^k)^2 = S^k.
\end{align*}

From the definition of $S^k$ and $R^k$, we have $\sqrt{S^k}=\|\bfTheta^k-\1(\btheta^k)^{\top}\|_F$ and $\sqrt{R^k}=\|\btheta^k-\theta^*\|$. 

We can decompose the quantity of interest using \eqref{eq:T_def}
\begin{align*}
    \langle\btheta^k-\theta^*,\sum_{i=1}^{N}\left(F_i(X_i^k,\theta_i^k)-\bar{F}_i(\btheta^k)\right)\rangle&=\langle\btheta^k-\btheta^{k-\tau_k},\sum_{i=1}^{N}\left(F_i(X_i^k,\theta_i^k)-\bar{F}_i(\btheta^k)\right)\rangle\notag\\
    &\hspace{20pt}+\langle\btheta^{k-\tau_k}-\theta^*,\sum_{i=1}^{N}\left(F_i(X_i^k,\theta_i^k)-\bar{F}_i(\btheta^k)\right)\rangle\notag\\
    &=\Tcal_1^k+\Tcal_2^k+\Tcal_3^k+\Tcal_4^k+\Tcal_5^k.
\end{align*}

We define $\Fcal_k=\{\mX^0,...,\mX^k,\bfTheta^0,...,\bfTheta^k\}$. Recall the definition of the TV distance: for any two probability measures $a,b$ on a set $E$
\begin{align*}
    d_{T V}(a, b)=\frac{1}{2} \sup_{f: E \rightarrow[-1,1]}\Big|\int f da-\int f db\Big|.
\end{align*}
This implies that
\begin{align*}
    \mathbb{E}\left[\Tcal_2^k\mid\Fcal_{k-\tau_k} \right]&= \left\langle\btheta^{k-\tau_k}-\theta^*,\vphantom{\mathbb{E}\left[\sum_{i=1}^{N}\left(F_i(X_i^k,\theta_i^{k-\tau_k})-\bar{F}_i(\theta_i^{k-\tau_k})\right)\mid\Fcal_{k-\tau_k}\right]}\mathbb{E}\left[\sum_{i=1}^{N}\left(F_i(X_i^k,\theta_i^{k-\tau_k})-\bar{F}_i(\theta_i^{k-\tau_k})\right)\mid\Fcal_{k-\tau_k}\right]\right\rangle\\
    &\leq\sum_{i=1}^{N}\sqrt{R^{k-\tau_k}}\left\|\mathbb{E}\left[F_i(X_i^k,\theta_i^{k-\tau_k})\mid\Fcal_{k-\tau_k}\right]-\bar{F}_i(\theta_i^{k-\tau_k})\right\|\notag\\
    &\leq B\sum_{i=1}^{N}\sqrt{R^{k-\tau_k}}\left(\left\|\theta_i^{k-\tau_k}\right\|+1\right) d_{TV}(P(X_i^{k}=\cdot),\mu_i).
\end{align*}


Using the geometric mixing time and $k\geq\tau_k$, we have 
$d_{TV}(P(X_i^{k}=\cdot),\mu_i)\leq\epsilon_k$.
This yields
\begin{align*}
    &\mathbb{E}\left[\Tcal_2^k\mid\Fcal_{k-\tau_k} \right]\notag\\
    &\leq B\epsilon_k\sum_{i=1}^{N}\left(\left\|\btheta^{k-\tau_k}-\btheta^{k}\right\|+\sqrt{R^k}\right)\left(r_i^{k-\tau_k}+\left\|\btheta^{k}-\btheta^{k-\tau_k}\right\|+\left\|\btheta^{k}\right\|+1\right)\notag\\
    &\leq B\epsilon_k\sum_{i=1}^{N}\left(\frac{1}{2N}\|\btheta^k\|+\frac{1}{2N}\sqrt{S^{k-\tau_k}}+\frac{1}{2N}+\sqrt{R^k}\right)\left(r_i^{k-\tau_k}+\frac{3}{2}\|\btheta^k\|+\frac{1}{2N}\sqrt{S^{k-\tau_k}}+\frac{3}{2}\right)\notag\\
    &\leq \frac{B\epsilon_k}{4}\sum_{i=1}^{N}\left(3\sqrt{R^k}+\frac{1}{N}\sqrt{S^{k-\tau_k}}+\frac{1+\|\theta^*\|}{N}\right)\Big(2r_i^{k-\tau_k}+3\sqrt{R^k}+3\|\theta^*\|+\frac{1}{N}\sqrt{S^{k-\tau_k}}+3\Big)\notag\\
    &\leq \frac{B\epsilon_k}{4}\sum_{i=1}^{N}\Big(3\sqrt{R^k}+3\|\theta^*\|+2r_i^{k-\tau_k}+\frac{1}{N}\sqrt{S^{k-\tau_k}}+3\Big)^2\notag\\
    &\leq \frac{5B\epsilon_k N}{4} \left(9R^k+9\|\theta^*\|^2+\frac{1}{N^2}S^{k-\tau_k}+9\right)+\frac{5B\epsilon_k}{4}\sum_{i=1}^{N}(r_i^{k-\tau_k})^2\notag\\
    &= \frac{5B\epsilon_k N}{4} \left(9R^k+9\|\theta^*\|^2+\frac{2}{N}S^{k-\tau_k}+9\right),
\end{align*}
where the second inequality is from \eqref{lem:theta_k+theta_k-tau:res1} of Lemma \ref{lem:theta_k+theta_k-tau}, and the first and the third inequalities use the triangular inequality.

The rest of the terms in \eqref{eq:T_def} are bounded by Lemma \ref{lem:innerproduct_term1} and \ref{lem:innerproduct_term345}. Putting the terms together,
\begin{align*}
    &\mathbb{E}[\langle\btheta^k-\theta^*,\sum_{i=1}^{N}\left(F_i(X_i^k,\theta_i^k)-\bar{F}_i(\btheta^k)\right)\rangle]\\
    &\leq \mathbb{E}[\Tcal_1^k+\Tcal_2^k+\Tcal_3^k+\Tcal_4^k+\Tcal_5^k]\\
    &\leq 30\epsilon_{k-\tau_k}\tau_k B^2\Big(N\mathbb{E}[R^k+NS^{k-\tau_k}+S^k+N \|\theta^*\|^2+N]\Big)\\
    &\hspace{5pt}+\frac{5B\epsilon_{k-\tau_k} N}{4} \left(9\mathbb{E}[R^k+9\|\theta^*\|^2+\frac{2}{N}S^{k-\tau_k}]+9\right)+ 48\epsilon_{k-\tau_k}\tau_k N BL\mathbb{E}[R^k]\\
    &\hspace{5pt}+\frac{\alpha}{2}\mathbb{E}[R^k]+\frac{45\epsilon_{k-\tau_k}\tau_k NBL}{2}(2\mathbb{E}[\|\theta^*\|^2]+1)+(\frac{NL^2}{\alpha}+\frac{L}{2})\mathbb{E}[S^{k}]+(\frac{4NL^2}{\alpha}+5L)\mathbb{E}[S^{k-\tau_k}]\\
    &\leq \epsilon_{k-\tau_k}\tau_k NB(\frac{45}{4}+30B+48L)\mathbb{E}[R^k]+\frac{\alpha}{2}\mathbb{E}[R^k]+(30\epsilon_{k-\tau_k}\tau_k B^2+\frac{NL^2}{\alpha}+\frac{L}{2})\mathbb{E}[S^k]\\
    &\hspace{5pt}+(30NB^2\epsilon_{k-\tau_k}\tau_k+\frac{5B\epsilon_{k-\tau_k}}{2}+\frac{4NL^2}{\alpha}+5L)\mathbb{E}[S^{k-\tau_k}]+\epsilon_{k-\tau_k}\tau_k N B(30B\hspace{-2pt}+\hspace{-2pt}\frac{45}{4}\hspace{-2pt}+\hspace{-2pt}45L)(\|\theta^*\|^2\hspace{-2pt}+\hspace{-2pt}1)\\
    &\leq \left(\frac{45B}{4}+30 B^2+48BL\right)N\epsilon_{k-\tau_k}\tau_k\mathbb{E}[R^k]+\frac{\alpha}{2}\mathbb{E}[R^k]+\left(\frac{5B}{N}+\frac{NL^2}{\alpha}+\frac{L}{2}\right)\mathbb{E}[S^k]\\
    &\hspace{5pt}+\left(5B+\frac{5}{12N}+\frac{4NL^2}{\alpha}+5L\right)\mathbb{E}[S^{k-\tau_k}]+N\epsilon_{k-\tau_k}\tau_k(30B^2+\frac{45B}{4}+45BL)\left(\|\theta^*\|^2+1\right),
\end{align*}
where the third inequality simply re-groups the terms, and the last inequality follows from $\epsilon_{k-\tau_k}\tau_k\leq\frac{1}{6NB}$.

\qed

\subsection{Proof of Lemma \ref{lem:innerproduct_term1}}


From the Cauchy-Schwarz inequality,
\begin{align*}
    T_1^k&\leq\sum_{i=1}^{N}\|\btheta^k-\btheta^{k-\tau_k}\|\|F_i(X_i^k,\theta_i^k)-\bar{F}_i(\btheta^k)\|\notag\\
    &\leq B\sum_{i=1}^{N}\|\btheta^k-\btheta^{k-\tau_k}\|\left(\|\theta_i^k\|+\|\btheta^k\|+2\right)\notag\\
    &\leq 3\epsilon_{k-\tau_k} \tau_k B^2\left(\|\btheta^k\|\hspace{-2pt}+\hspace{-2pt}\sqrt{S^{k-\tau_k}}\hspace{-2pt}+\hspace{-2pt}1\right)\sum_{i=1}^{N}\left(r_i^k+2\|\btheta^k\|\hspace{-1pt}+\hspace{-1pt}2\right)\notag\\
    &\leq 6\epsilon_{k-\tau_k}\tau_k B^2\sum_{i=1}^{N}\left(\|\btheta^k\|+\sqrt{S^{k-\tau_k}}+r_{i}^{k}+1\right)^2\notag\\
    &\leq 6\epsilon_{k-\tau_k}\tau_k B^2\sum_{i=1}^{N}\left(\sqrt{R^k}+\|\theta^*\|+\sqrt{S^{k-\tau_k}}+r_{i}^{k}+1\right)^2\notag\\
    &\leq 30\epsilon_{k-\tau_k}\tau_k B^2\sum_{i=1}^{N}\left(R^k+S^{k-\tau_k}+(r_{i}^{k})^2+\|\theta^*\|^2+1\right)\notag\\
    &= 30\epsilon_{k-\tau_k}\tau_k B^2\left(NR^k+NS^{k-\tau_k}+S^k+N (\|\theta^*\|^2+1)\right),
\end{align*}
where the second inequality results from Lemma \ref{lem:B_bounded}, and the third and fifth inequalities use the triangular inequality. The third inequality also uses \eqref{lem:theta_k+theta_k-tau:res1} of Lemma \ref{lem:theta_k+theta_k-tau}.

\qed

\subsection{Proof of Lemma \ref{lem:innerproduct_term345}}

Due to the Lipschitz continuity in Assumption~\ref{assump:F_Lipschitz} and the Cauchy-Schwarz inequality,
\begin{align*}
    &\Tcal_3^k+\Tcal_4^k+\Tcal_5^k\\
    &\leq\sqrt{R^{k-\tau_k}}\sum_{i=1}^{N}L\left(r_i^k+r_i^{k-\tau_k}+\|\btheta^{k-\tau_k}-\btheta^{k}\|\right)\notag\\
    &\leq L\left(\|\btheta^{k-\tau_k}-\btheta^{k}\|+\sqrt{R^k}\right)\sum_{i=1}^{N}\left(r_i^k+2r_i^{k-\tau_k}+2\|\btheta^{k-\tau_k}-\btheta^{k}\|\right)\notag\\
    &\leq L\sum_{i=1}^{N}\Big( 3\epsilon_{k-\tau_k}\tau_k B\|\btheta^k\|+3\epsilon_{k-\tau_k}\tau_k B \sqrt{S^{k-\tau_k}}+3\epsilon_{k-\tau_k}\tau_kB+\|\btheta^{k}-\theta^*\|\Big)\notag\\
    &\hspace{20pt}\times\Big(r_i^k+2r_i^{k-\tau_k}+6\epsilon_{k-\tau_k} \tau_k B\|\btheta^k\|+6\epsilon_{k-\tau_k} \tau_k B \sqrt{S^{k-\tau_k}}+6\epsilon_{k-\tau_k}\tau_k B\Big)\notag\\
    &=3\epsilon_{k-\tau_k}\tau_k BL\sum_{i=1}^{N}(\|\btheta^k\|+\sqrt{S^{k-\tau_k}}+1)(r_i^k+2r_i^{k-\tau_k})+2L(3\epsilon_{k-\tau_k}\tau_k B)^2\sum_{i=1}^{N}\left(\|\btheta^k\|+\sqrt{S^{k-\tau_k}}+1\right)^2\notag\\
    &\hspace{20pt}+L\sum_{i=1}^{N}\|\btheta^k-\theta^*\|\left(r_i^k+2r_i^{k-\tau_k}\right)+6\epsilon_{k-\tau_k}\tau_k BL\sum_{i=1}^{N}\|\btheta^k-\theta^*\|\Big(\|\btheta^k\|+\sqrt{S^{k-\tau_k}}+1\Big),
\end{align*}
where the second inequality uses the triangular inequality, and the third inequality follows from \eqref{lem:theta_k+theta_k-tau:res1} of Lemma \ref{lem:theta_k+theta_k-tau}. The last inequality simply re-groups the terms. Then, using the Cauchy-Schwarz inequality and the fact that $a\cdot b\leq\frac{c}{2}a^2+\frac{1}{2c}b^2$ for any scalars $a,b,c$ ($c>0$), we can further simplify
\begin{align*}
    &\Tcal_3^k+\Tcal_4^k+\Tcal_5^k\\
    &\leq \frac{3\epsilon_{k-\tau_k}\tau_k BL}{2}\hspace{-3pt}\sum_{i=1}^{N}\hspace{-3pt}\left(\hspace{-2pt}(\|\btheta^k\|\hspace{-3pt}+\hspace{-3pt}\sqrt{S^{k-\tau_k}}+1)^2\hspace{-3pt}+\hspace{-3pt}(r_i^k+2r_i^{k-\tau_k})^2\hspace{-2pt}\right)+18NL\epsilon_{k-\tau_k}^2 \tau_k^2 B^2(\|\btheta^k\|+\sqrt{S^{k-\tau_k}}+1)^2\notag\\
    &\hspace{20pt}+\frac{\alpha}{2}R^k+\frac{L^2}{2\alpha}\left(\sum_{i=1}^{N}(r_i^k+2r_i^{k-\tau_k})\right)^2+3\epsilon_{k-\tau_k}\tau_k N BL\left(R^k+(\|\btheta^k\|+\sqrt{S^{k-\tau_k}}+1)^2\right)\notag\\
    &\leq \frac{9\epsilon_{k-\tau_k}\tau_kN BL}{2}(\|\btheta^k\|^2+S^{k-\tau_k}+1)+3\epsilon_{k-\tau_k}\tau_k BL\sum_{i=1}^{N}((r_i^k)^2+4(r_i^{k-\tau_k})^2)\notag\\
    &\hspace{20pt}+9\epsilon_{k-\tau_k}\tau_k BL(\|\btheta^k\|^2+S^{k-\tau_k}+1)+\frac{\alpha}{2}R^k+\frac{N L^2}{\alpha}\sum_{i=1}^{N}\left((r_i^k)^2+4(r_i^{k-\tau_k})^2\right)\notag\\
    &\hspace{20pt}+3\epsilon_{k-\tau_k}\tau_k N BL R^k+9\epsilon_{k-\tau_k}\tau_k N BL(\|\btheta^k\|^2+S^{k-\tau_k}+1)\notag\\
    &\leq \frac{45\epsilon_{k-\tau_k}\tau_k N BL}{2}(\|\btheta^k\|^2+S^{k-\tau_k}+1)+\frac{\alpha}{2}R^k+3\epsilon_{k-\tau_k}\tau_k N BLR^k+\frac{NL^2}{\alpha}\left(S^{k}+4S^{k-\tau_k}\right)\notag\\
    &\hspace{20pt}+3\epsilon_{k-\tau_k}\tau_k NBL\left(S^{k}+4S^{k-\tau_k}\right)\notag\\
    &\leq 48\epsilon_{k-\tau_k}\tau_k N BLR^k+\frac{\alpha}{2}R^k+(\frac{NL^2}{\alpha}+\frac{L}{2})S^{k}+(\frac{4NL^2}{\alpha}\hspace{-2pt}+\hspace{-2pt}5L)S^{k-\tau_k}+\frac{45\epsilon_{k-\tau_k}\tau_k NBL}{2}(2\|\theta^*\|^2\hspace{-2pt}+\hspace{-2pt}1),
\end{align*}
where the last two inequality simply re-group the terms and simplify the expressions using the conditions on the step size.
\qed

\end{document}